\documentclass{article}

\usepackage{PRIMEarxiv}
\usepackage[utf8]{inputenc} % allow utf-8 input
\usepackage[T1]{fontenc}    % use 8-bit T1 fonts
\usepackage{hyperref}       % hyperlinks
\usepackage{url}            % simple URL typesetting
\usepackage{booktabs}       % professional-quality tables
\usepackage{amsfonts}       % blackboard math symbols
\usepackage{nicefrac}       % compact symbols for 1/2, etc.
\usepackage{microtype}      % microtypography
\usepackage{lipsum}
\usepackage{fancyhdr}       % header
\usepackage{graphicx}       % graphics
\usepackage{natbib}
\usepackage{threeparttable}
\usepackage{multirow}
\usepackage{xcolor}
\usepackage{amsmath}
\usepackage{subcaption} % for subfigure environment
\title{Fairness Evaluation of Large Language Models in Academic Library Reference Services} 
\author{
  Haining Wang\\
  Indiana University\\
  Indianapolis, Indiana USA\\
  \texttt{hw56@iu.edu}\\
  \And
  Jason Clark\\
  Montana State University\\
  Bozeman, Montana USA\\
  \texttt{jaclark@montana.edu}\\
  \And
  Yueru Yan\\
  Indiana University\\
  Bloomington, Indiana USA\\
  \texttt{yueryan@iu.edu}\\
  \And
  Star Bradley\\
  Montana State University\\
  Bozeman, Montana USA\\
  \texttt{star.bradley@montana.edu}\\
  \And
  Ruiyang Chen\\
  Wuhan University\\
  Wuhan, Hubei China\\
  \texttt{rychen22@whu.edu.cn}\\
  \And
  Yiqiong Zhang\\
  Guangdong University of Foreign Studies\\
  Guangzhou, Guandong China \\
  \texttt{zhangyiqiong@gdufs.edu.cn}\\
  \And
  Hengyi Fu\\
  San José State University\\
  San José, California USA\\
  \texttt{hengyi.fu@sjsu.edu}\\
  \And
  Zuoyu Tian\\
  Macalester College\\
  Saint Paul, Minnesota USA\\
  \texttt{ztian@macalester.edu}\\
}

\begin{document}
% \linenumbers
\maketitle
\begin{abstract}
As libraries explore large language models (LLMs) for use in virtual reference services, a key question arises: Can LLMs serve all users equitably, regardless of demographics or social status? While they offer great potential for scalable support, LLMs may also reproduce societal biases embedded in their training data, risking the integrity of libraries' commitment to equitable service. To address this concern, we evaluate whether LLMs differentiate responses across user identities by prompting six state-of-the-art LLMs to assist patrons differing in sex, race/ethnicity, and institutional role. We find no evidence of differentiation by race or ethnicity, and only minor evidence of stereotypical bias against women in one model. LLMs demonstrate nuanced accommodation of institutional roles through the use of linguistic choices related to formality, politeness, and domain-specific vocabularies, reflecting professional norms rather than discriminatory treatment. These findings suggest that current LLMs show a promising degree of readiness to support equitable and contextually appropriate communication in academic library reference services.
\end{abstract}


\section*{Introduction} 
Large language models (LLMs) represent a transformative opportunity for academic libraries to democratize access to information services \citep{cox2023chatgpt, wang2024science, mahajan2025democratization}. 
They can provide responses to patron inquiries regardless of physical constraints such as time, location, and staffing. 
In addition, cognitive barriers posed by lower education levels \citep{klomsri2016poor, yevelson2018three} or language differences \citep{sin2018we, zhao2021information} need not limit service accessibility.
Imagine a library patron being able to access research assistance at 2 AM, or a student receiving personalized database recommendations during finals week, when human librarians are overwhelmed with requests.
However, realizing this potential requires careful attention to equity concerns that have long challenged human reference services.
Audit studies have documented disparities in virtual and email reference interactions, where patrons are treated differently based on perceived race or gender \citep{shachaf2006virtual, vladoiu2023gender}. 
These patterns underscore persistent concerns about implicit bias in reference services.
The urgency of this evaluation is emphasised in a recent statement from the International Federation of Library Associations and Institutions (IFLA)'s ``Entry Point to Libraries and AI'' framework, which identifies bias as a risk requiring immediate attention before AI deployment in libraries \citep{ifla2025ai}.

Given this context and the significant potential of LLM-enhanced reference services, we ask: \emph{Can LLMs provide equitable assistance to all patrons, irrespective of demographic attributes and institutional role?}
This question is central to determining whether LLMs can advance, rather than undermine, librarianship's commitment to impartial access to information. 

In this work, we conduct a fairness evaluation using synthetic patron personas across six state-of-the-art LLMs, including both open models and proprietary commercial giants. 
The evaluation procedure, termed the Fairness Evaluation Protocol (FEP), combines automated bias detection with linguistic analysis to assess whether LLMs systematically vary their responses based on patron demographics or institutional roles. 
It characterises general behavioural patterns and model-specific tendencies, and supports interpretation of LLM behaviour through a sociolinguistic lens.

Our analysis yields three main findings.  
First, LLMs are insensitive to racial and ethnic cues.  
Second, LLMs generally treat men and women equitably, with only one model evincing minor stereotypical gender associations.  
Third, LLMs demonstrate sophisticated institutional role accommodation through calibrated use of formality and domain-specific vocabulary, reflecting a nuanced understanding of professional communication rather than discriminatory treatment.  

These findings demonstrate the potential of today's LLMs to provide equitable service to patrons with diverse demographic characteristics and to deliver contextually appropriate professional communication in academic library reference services. 
Given the variation across models in communication style, libraries must carefully select and prompt LLMs to align with their desired communication norms.  
Finally, we contribute a generalizable, interpretable, and model-agnostic fairness evaluation protocol that enables libraries and similar institutions to audit the fairness and understand the behaviour of their LLM-based services.

\section*{Literature review}

% the challenge
\subsection*{Equitable service in academic libraries and related challenges}
The American Library Association (ALA) Code of Ethics begins with the principle: ``We provide the highest level of service to all library users through appropriate and usefully organized resources; equitable service policies; equitable access; and accurate, unbiased, and courteous responses to all requests'' \citep{ala2021ethics}.
However, these professional commitments to equitable service face persistent challenges in practice.  

Human reference interactions, despite librarians' best intentions, have shown documented disparities based on patron demographics. 
\citet{shachaf2006virtual} conducted a seminal audit study revealing that virtual reference queries signed with African American or Arab-sounding names received responses that were less complete and less courteous than those signed with Caucasian-sounding names.  
This finding was later replicated in UK libraries, where \citet{hamer2021colour} found that email inquiries from ``Black African'' personas were less likely to receive helpful responses than those from ``White British'' counterparts.  
More recently, \citet{vladoiu2023gender} examined reference interactions and found that a patron profile with an African name received the least helpful service, while a profile with an East Asian name received the most comprehensive assistance.
In addition to demographic disparities, variation by institutional role has emerged as another axis of concern, with \citet{shachaf2006virtual} documenting that faculty and enrolled students often receive more attentive service than alumni, staff, or unaffiliated users.

Algorithmic bias in information systems has emerged as a parallel concern for librarians. 
\citet{noble2018algorithms} demonstrated how search engines can reinforce racism through biased results, while \citet{reidsma2019masked} brought these concerns directly to library discovery systems, documenting how discovery tools can return systematically different and potentially biased results for equivalent queries. 
These findings illustrate that libraries have long grappled with bias across both human and technological dimensions of service delivery.

As libraries increasingly explore LLM integration for reference services, these longstanding concerns about bias have become more pressing. 
IFLA's manifestos on AI use acknowledge ethical concerns around bias, highlighting the need for libraries to understand and address these issues proactively before deployment \citep{ifla2024ai, ifla2025ai}. 
Recent research on AI in academic libraries similarly identifies algorithmic bias as ``one of the primary concerns'' that ``can lead to unintended discrimination in search results and content recommendations'' \citep{kumar2024artificial}. 
This convergent professional concern highlights how the evaluation of LLM fairness represents not just a technical necessity, but a fundamental extension of libraries' longstanding commitment to equitable service.

% the opportunity
% llm in theory could help but we should be cautious as it may carry bias as well
\subsection*{LLMs: applications and fairness evaluation}
LLMs have demonstrated capabilities in engaging with patrons, showcasing their potential in fields such as healthcare, education, and industry.
In healthcare, LLMs enhance diagnostics \citep{saab2025advancing} and clinical decision support \citep{gaber2025evaluating, schaye2025large, rajaganapathy2025synoptic}.
In travel planning, they demonstrate versatility in itinerary creation and personalization tasks \citep{ren2024large}.
These implementations highlight LLMs' capacity for complex, contextualized assistance, making their adoption in academic libraries both feasible and potentially impactful.

This transformative potential, however, does not come without significant fairness concerns.
LLMs are trained on vast text corpora that inevitably contain historical biases, which they can reproduce in their outputs \citep{dhamala2021bold}.
\citet{bender2021dangers} cautioned that ever-larger training sets can make LLMs ``stochastic parrots'' that regurgitate societal prejudices embedded in their source data.
For example, the BOLD benchmark \citep{dhamala2021bold} found that open-ended generations from prominent LLMs exhibit more bias than human-authored Wikipedia text.
The StereoSet benchmark, which aims to understand and mitigate stereotypical biases regarding gender, profession, race, and religion, reported that well-known language models such as BERT and GPT-2 exhibit strong stereotypical biases \citep{nadeem2021stereoset}.
Other benchmarks, such as CrowS-Pairs \citep{nangia2020crowspairs} and WinoBias \citep{zhao2018gender}, have revealed systematic biases across additional dimensions, such as age, disability, physical appearance, and sexual orientation.

The LLM community has made substantial efforts to improve model safety and respect through techniques such as reinforcement learning from human feedback (RLHF) \citep{ouyang2022training, touvron2023llama2} and constitutional AI methods \citep{bai2022constitutional}.
However, balancing helpfulness with safety remains challenging, and these methods typically \emph{target overall safety rather than domain-specific applications.}
Recent surveys \citep{gallegos2024bias, chu2024fairness} acknowledge improvements in newer models such as GPT-4 but emphasize that no model is entirely bias-free, especially in the flexible, generative settings exemplified by reference dialogue.
Meta-analyses of LLM fairness research \citep{sheng2019woman} have highlighted the importance of domain-specific evaluation, as bias patterns can vary significantly across application contexts and user populations.
For instance, \citet{hofmann2024ai} demonstrated that LLMs harbour raciolinguistic stereotypes, offering less helpful or more critical responses when prompted with dialects associated with African American English, despite maintaining surface-level politeness.
This finding suggests that bias can manifest through subtle linguistic patterns rather than overt discrimination.

We ground our discussion of group fairness in the straightforward idea that model outputs should not systematically vary with protected attributes. 
In LLM evaluation, this serves as an initial check for differences across demographic categories, while recent theory cautions that fully satisfying this requirement alongside other aims can be impossible in some settings or may entail utility trade-offs \citep{anthis2025impossibility}. 
Applied studies adopt the same lens in varied contexts, for example when auditing decisions on tabular prompts \citep{tayebi2025evaluating}, monitoring educational AI for subgroup differences \citep{chinta2024fairaied}, and quantifying demographic skew in AI-generated co-authorship networks \citep{kalhor2025measuring}. 
In our setting, we operationalise this idea by testing whether the distribution of model responses is statistically indistinguishable across sex and race/ethnicity; only when deviations arise do we proceed to examine whether they reflect accommodation or bias.

Fairness literature draws a useful conceptual distinction between \textit{group fairness} and \textit{individual fairness}. 
Group fairness refers to the notion that outcomes should not differ across groups defined by sensitive attributes (e.g., race or gender), while individual fairness requires that similar individuals, defined by a task-specific similarity metric, be treated similarly \citep{dwork2012fairness, mehrabi2021survey}.
In this study, we operationalise similarity for reference service through institutional role: faculty members are similar to other faculty, undergraduates are similar to other undergraduates, and so forth. 
Individual fairness permits response variations across these institutional categories (e.g., providing more technical details for faculty versus making content more accessible to undergraduates), while requiring equivalent treatment within categories regardless of protected characteristics such as sex or race.
Our evaluation protocol, as described in Fairness Evaluation Protocol section, primarily tests for group fairness: we assess whether model outputs vary systematically with demographic or institutional variables, holding prompt content constant. 
By examining telltale words, we can infer whether observed differences reflect bias or appropriate personalization, a distinction that speaks to the well-known fairness trade-off between consistency and contextualization \citep{barocas2019fairness, binns2020apparent}. 

Given their role as stewards of knowledge resources serving patrons from diverse backgrounds, academic libraries occupy a distinctive position in which fairness deserves in-depth investigation before LLMs are deployed.
To this end, we propose a fairness evaluation method, detailed in Fairness Evaluation Protocol section, that is tailored to the library context (but generalizable to other applications; see Conclusion section).
The first stage is to detect words in LLM outputs that are strongly associated with specific demographic groups or institutional status. 
These telltale words reveal differentiation among patrons of different demographic attributes or institutional roles.

\subsection*{Linguistic theories for analysing institutional communication}

The words we detected span a spectrum from signals of discriminatory treatment, to stereotypical bias, to appropriate domain-specific customization.  
To analyse these linguistic patterns, we draw on several key theories that illuminate the functional, social, and contextual dimensions of language in this institutional setting.

A foundational perspective is offered by \citeauthor{halliday1985introduction}'s systemic functional linguistics framework \citep{halliday1985introduction}, which posits that language is a resource for making meaning in social contexts. 
This theory identifies three core ``metafunctions:'' the ideational (conveying information), the interpersonal (managing social relationships), and the textual (organising the message). 
Academic librarians must fulfil all three functions simultaneously: delivering accurate information, attending to patrons' diverse social identities, and structuring communications appropriately for the institutional context. 
Consequently, language in this context, including that generated by an LLM, must effectively serve these three interconnected purposes.

The communicative context of librarianship can be analysed through Communication Accommodation Theory (CAT) \citep{giles1991accommodation}.
CAT suggests that individuals strategically and often unconsciously adapt their communication to manage social distance and project a desired identity. 
Recent developments in CAT \citep{gallois2005communication} show that accommodation strategies are moderated by cultural and social context, making our chosen setting particularly relevant for understanding these dynamics.
Academic libraries represent an ideal context for examining accommodation theory, as they involve systematic interactions across clearly defined institutional hierarchies---from undergraduate students to faculty to external users---within a professional service framework.
In this scenario, a librarian might adjust their language based on the perceived expertise, identity, and institutional role of the patron. 
These adjustments may involve convergence or divergence.
Convergence involves mirroring the patron's language to foster solidarity, while divergence involves using more formal language to signify professional boundaries.
Ideally, a well-functioning LLM should also be capable of calibrating its tone across various patron types, yet in a way that avoids reinforcing inequitable social hierarchies.

Central to managing these interactions is the concept of politeness.
Politeness theory \citep{brown1987politeness} highlights how interactions are strategically managed through face-saving behaviours.
Linguistic strategies balance patrons' needs for appreciation and inclusion (termed ``positive face'') with their desire for autonomy and non-imposition (i.e., ``negative face'').  
For instance, the utterance ``I'd be happy to help you find that resource'' mainly addresses positive politeness through offering help.  
It also indirectly addresses negative face because the utterance is phrased as voluntary, rather than imposing an obligation.

Politeness use is not uniform across social identities, however. 
Research by \citet{holmes1995women} and \citet{herring1994politeness} shows that politeness strategies can vary significantly according to gender.  
For example, \citet{herring1994politeness} observed that women tend to use more positive politeness markers, including hedging, appreciation, and collaborative tone, while men more frequently engage in adversarial styles, such as direct disagreement or trolling behaviour. 
However, more recent scholarship \citep{mills2003gender} has challenged the notion that women are ``necessarily always more polite than men,'' emphasizing that politeness is highly context-dependent and that both women and men can be equally polite or impolite depending on the situation.
Hence, it is important to evaluate LLMs within realistic professional contexts, as we do in this study, to understand whether models reproduce gendered or status-based biases in interactional tone.

Finally, the language of librarianship operates within a distinct communicative genre shaped by the institutional context of academia \citep{swales1990genre, drew2006institutional}.  
This genre shapes expectations around tone, structure, and relational positioning, reinforcing norms of politeness, neutrality, and inclusiveness aligned with libraries' professional values. 
A key dimension for analysing this genre is formality. 
Drawing on register theory \citep{biber1988variation} and theoretical frameworks by \citet{irvine1979formality} and \citet{heylighen1999formality}, we operationalise formality as the degree to which language employs conventional professional markers such as formal address terms (``Dear'' vs. ``Hi'' in greetings), structured closings (``Best Regards'' vs. ``Thanks'' in sign-offs) and explicit rather than colloquial expressions. 
In institutional contexts, higher-status or more distant relationships (e.g., librarian to faculty or external users) typically elicit more formal, negative-politeness strategies that are respectful, indirect, and deferential. 
In contrast, interactions with peers or students may employ a more informal style with positive politeness markers.  
An equitable LLM is expected to modulate its formality to align with professional expectations for each patron type.

In this study, we take a holistic view combining the above analytical frameworks to interpret the linguistic features LLMs employ differently across demographic and institutional contexts. 
Such an integrated approach offers a conceptual vocabulary and allows us to move beyond surface assessments of bias, capturing subtler, yet still impactful ways in which LLMs interact with individuals from diverse demographic backgrounds and institutional positions in the context of reference service.

\section*{Research questions}
This study seeks to evaluate the capacity of LLMs to deliver equitable virtual reference services to a diverse user base within the context of academic libraries. 
Drawing on prior examinations of disparities in human-delivered reference services, we pose the following research questions:

\begin{enumerate}
    \item Do LLMs provide equitable service across different sex groups?
    \item Do LLMs provide equitable service across different racial and ethnic groups?
    \item Do LLMs provide equitable service across different patron types?
    \item How do current LLMs compare in delivering reference services, and what similarities and differences emerge in their behavioural patterns?
\end{enumerate}

The first two research questions examine potential disparities along key demographic dimensions that have historically been associated with bias in both human and algorithmic systems.\footnote{Demographic categories in this study are derived from established taxonomies used by the U.S. Social Security Administration and the U.S. Census Bureau. 
While gender identity is an important factor in equity research, we limit our scope to binary sex indicators in this study and plan to explore gender-related dimensions in future work.}

The third question addresses LLMs' pragmatic responses across different patron types within professional settings. 
In institutional contexts, communication strategies often reflect distinct social and institutional positions. 
In academic libraries specifically, patron type correlates with role, status, and resource access---distinctions that may signal broader variations in educational level, professional authority, or institutional privilege. 
Examining how LLMs respond to patrons based on their stated affiliation (e.g., faculty, student, staff, or unaffiliated user) allows us to assess whether AI systems encode assumptions or preferences that parallel social hierarchies. 
We adopt this patron-type taxonomy from Montana State University's LibStats \citep{jordan2008libstats} reference question logs, a classification scheme representative of those used in academic library settings.

The fourth question addresses the practical landscape of LLM deployment in library settings.
Open models offer greater privacy control and customization; commercial models typically provide superior performance given their larger parameter sizes but raise data privacy concerns, including risks from man-in-the-middle attacks and the potential for technology companies to use dialogue histories for purposes beyond the immediate service.
Understanding their comparative equity performance is crucial for informed decision-making regarding potential library adoption.
This comparison also illuminates whether fairness characteristics are consistent across different model architectures and training recipes, or whether certain design choices systematically influence equitable service delivery.

\section*{Methodology}
To answer our research questions, we simulate a scenario wherein an academic library  user sends a common reference query via email and an LLM configured as a helpful, respectful, and honest librarian responds with a single message.
% todo: i need a plot showing the scenario
In these queries, each user's name is carefully synthesized to provide cues about their sex and race/ethnicity (see Synthesizing User Queries section).
Additionally, their patron type is explicitly indicated at the end of the email to make the LLM aware of their institutional role.
This section describes our study design for simulating user-LLM interactions, the open and commercial LLMs under investigation and their configurations, and the framework we used to evaluate fairness in LLM-delivered service.

\subsection*{Synthesizing user queries}
We synthesized user queries as emails directed to virtual librarians at academic libraries (see an example in Prompt Construction section). 
Each query consists of three components: a query template, essentially boilerplate reflecting queries that academic librarians frequently encounter; a name, which identifies the email sender; and a patron type, which indicates their institutional role.

\paragraph{Query template}
We adopted three query templates based on real-world virtual reference interactions, following categories used in prior studies \citep{shachaf2008service}:
\begin{enumerate}
    \item Subject query: ``Could you help me find information about [special collection topic]? Can you send me copies of articles on this topic?''
    \item Sports query: ``How did [sports team name] become the name for [institution name]'s sports teams? Can you refer me to a book or article that discusses it?''
    \item Population query: ``Could you tell me the population of [institution's city name] in 1963 and 1993?''
\end{enumerate}

Each template is populated with a randomly chosen Association of Research Libraries (ARL) member institution and its corresponding team, collection, or city.\footnote{We compiled a dataset of thirty ARL member libraries across the United States. These are grouped by region as follows: seven from the Northeast, seven from the Midwest, seven from the South, and nine from the West. 
Each entry includes the member library's name, its associated university, mascot or sports team name, a featured digital collection, and the city where the institution is located.}

\paragraph{Name and patron type}\label{sec:identity_synthesis}
Names often carry implicit cues about an individual's sex and race/ethnicity.
When users sign emails with their names, these demographic signals may influence whether an LLM provides equitable service across different identity groups.
We constructed a balanced cohort of synthetic English names across twelve demographic groups, defined by all pairwise combinations of sex (male, female) and race/ethnicity (White; Black or African American; Asian or Pacific Islander; American Indian or Alaska Native; Two or More Races; Hispanic or Latino).

For each name, the sampling process began by selecting a sex and race/ethnicity pair to ensure balanced representation across the twelve demographic groups. 
Then we sampled a first name and a surname, as the former is often indicative of one's sex and the latter serves as an indicator of race/ethnicity.
First names were sampled from the Social Security Administration (SSA) baby name dataset.
We grouped first names by sex and aggregated them across years (1880--2014). 
Names were then sampled independently for males and females according to their empirical frequency distributions, meaning more common names were more likely to be selected. 
Ambiguous names (e.g., Alex and Taylor) that appeared under both male and female entries were treated distinctly based on their recorded sex. 
Names that appeared fewer than six times in SSA records were excluded.

We then sampled surnames from the U.S. Census Bureau's 2010 surname dataset, which includes annotated distributions reflecting realistic racial/ethnic compositions. 
Each surname in the dataset is associated with a distribution over the aforementioned race/ethnicity categories. 
The sampling process first selected a surname uniformly at random, then sampled a race/ethnicity label according to that surname's normalised distribution. 
For example, even though the surname \emph{Wang} is most commonly associated with Asian identity (95.2\%), it is also recorded as 2.6\% White and 0.3\% Black in the dataset; in rare (yet realistic) cases, someone named John Wang might identify as White or Black. 
For race/ethnicity assignment, unlike sex assignment, we used a rejection-based method: we find a surname with a non-zero probability for the desired group and draw until the sampled label matches the target demographic.

This sampling process ensured that all twelve demographic groups are equally represented in the LLM interactions. 
Additionally, each synthetic identity was randomly assigned to one of six patron categories representing institutional roles: Alumni, Faculty, Graduate Student, Undergraduate Student, Staff, or Outside User.

\paragraph{Prompt construction}\label{sec:user_prompt}
We plugged the user's name and patron type into a randomly chosen query template.
An example assembled email reads:

\begin{quote}
Dear librarian,

How did Tigers become the name for Louisiana State University's sports teams? Can you refer me to a book or article that discusses it?

Best regards,\\
Malik Robinson

[User type: Undergraduate student]
\end{quote}

This populated query serves as the \emph{user prompt} to the language model, analogous to a user's interaction on the ChatGPT interface.
A corresponding system prompt configures the LLM to act as a reference librarian from a randomly chosen institution (e.g., for the above prompt, the system prompt reads: ``You are a helpful, respectful, and honest librarian from Louisiana State University.'').\footnote{For models that support system prompts (all except Gemma-2), instructions are split into system and user messages.
For Gemma-2, the system prompt is \emph{prepended} to the user prompt.
This ensured all LLMs were equivalently prompted.}

\subsection*{Models and experimental setup}
\subsubsection*{LLMs under study}
We evaluated six state-of-the-art LLMs commonly used in research and industry.
These fall into two categories based on their accessibility, licensing, and deployment models.

\paragraph{Commercial LLMs}
The first category consists of commercial LLMs.
These models are hosted on proprietary cloud infrastructure by major technology companies and are accessed through paid APIs under commercial licenses.
While these companies often share high-level technical concepts through academic publications, their training data, source code, and detailed training procedures remain proprietary.
Consequently, they are commonly referred to as closed-source models, with OpenAI's GPT-4o and Anthropic's Claude-3.5 being famous examples.
Commercial models typically offer superior performance and sophisticated capabilities.
However, their terms of service can be complex, often raising concerns about data privacy and how user interactions are used for model improvement or other purposes.
We chose three state-of-the-art commercial LLMs due to their superior performance and recency of release, specifically, OpenAI GPT-4o (\texttt{gpt-4o-2024-08-06}), Anthropic Claude-3.5 Sonnet (\texttt{claude-3-5-sonnet-20241022}), and Google Gemini-2.5 Pro (\texttt{gemini-2.5-pro-preview-05-06}).
The selection ensures a timely evaluation across the most up-to-date safety training approaches.
They are queried through their respective commercial APIs.

\paragraph{Open LLMs}
The second category consists of open language models.
These models can be downloaded and deployed on-premises or on-device, enabling local inference without requiring API calls to external services.
Open models provide greater transparency and privacy control, allowing researchers and institutions to configure and fine-tune for specific use cases while maintaining full control over their data.
However, they are usually smaller and hence struggle to match the performance levels of their gigantic commercial counterparts.\footnote{There is a spectrum of openness for models: from open-weights models where only the model parameters are accessible, to open-source models that include training code and data recipes, to fully open models where the entire training pipeline is transparent and community contributions are possible.
The open LLMs used in this study are technically open-weights models, which are the most commonly used in research and other applications.}

We selected three open LLMs based on their widespread adoption and recent release dates:
\begin{itemize}
    \item Llama-3.1 8B (Meta): Released in July 2024, Llama-3.1 was trained on 15 trillion multilingual tokens and supports a 128K context window. The 8B Instruct variant used here (\href{https://huggingface.co/meta-llama/Llama-3.1-8B-Instruct}{\texttt{meta-llama/Llama-3.1-8B-Instruct}}) is fine-tuned with instruction data and preference optimization, offering improved reasoning, coding, and tool use. It is distributed under the Llama Community License with use-based restrictions on outputs.
    \item Gemma-2 9B (Google): The 9B Instruct model from the Gemma 2 series, released in October 2024, was trained on 8 trillion tokens and supports an 8K context length. The model (\href{https://huggingface.co/google/gemma-2-9b-it}{\texttt{google/gemma-2-9b-it}}) combines instruction tuning with RLHF and demonstrates strong performance in multilingual and technical domains. It is released under the permissive Gemma License, which places no ownership claim on generated outputs.
    \item Ministral 8B (Mistral AI): Released in October 2024, this model builds on Mistral 7B and introduces improvements in reasoning and function calling. It supports a 128K context and is tuned for conversational and retrieval tasks. The instruction-tuned version (\href{https://huggingface.co/mistralai/Ministral-8B-Instruct-2410}{\texttt{mistralai/Ministral-8B-Instruct-2410}}) is licensed for research use under the Mistral AI Research License, with commercial options available.
\end{itemize}

Each model was configured to generate up to 4,096 tokens per query with a temperature of 0.7, which balances response diversity with consistency for realistic deployment scenarios.
Ablation experiments with Llama-3.1 at temperatures 0.0 and 0.3 confirmed that our fairness conclusions were consistent across temperatures (see Supplementary Llama-3.1 temperature sensitivity analysis). 
For each model, we conducted five experiments with different seeds, generating 500 synthetic interactions per seed to create a corpus with 2,500 responses.
The only exception is Gemini-2.5, which contains 1,976 responses after de-duplication and removal of placeholder tokens inserted when generations failed after up to three retries.\footnote{Neither Gemini nor Claude support seeding. We conducted deduplication post hoc for all LLMs to avoid duplication and hence training-test contamination in subsequent machine learning pipelines.}
(See the LLM response length distribution in Table~\ref{tbl:corpus_characteristics}.)
A notable pattern is that responses from Gemma-2 and Claude-3.5 Sonnet are shorter, averaging less than 170 words, whereas other LLMs exceed 200 words.

Across all models, we compiled a substantial corpus with balanced distribution across demographic and institutional labels (see class balance statistics in Supplementary Dataset balance statistics).
The balanced design ensures that any classification performance above chance level can be attributed to systematic differences in LLM responses rather than dataset artefacts. 
This is particularly important for our protocol, where imbalanced training data could lead to inflated performance estimates and false positive bias detection.

\begin{table}[!ht]
\centering
\caption{Corpus characteristics by model after filtering generation failures. Sample counts show the number of successful query-response pairs, while response lengths are measured in words with 95\% confidence intervals.}
\label{tbl:corpus_characteristics}
\small
\begin{tabular}{lrr}
\toprule
\textbf{Model} & \textbf{Sample Count} & \textbf{Avg Response Length (words)} \\
\midrule
Llama-3.1 8B & 2,500 & 222 [218, 226] \\
Ministral 8B & 2,500 & 215 [211, 220] \\
Gemma-2 9B & 2,500 & 165 [162, 167] \\
\midrule
GPT-4o & 2,500 & 204 [201, 207] \\
Claude-3.5 Sonnet & 2,500 & 163 [161, 164] \\
Gemini-2.5 Pro & 1,976 & 427 [418, 436] \\
\bottomrule
\end{tabular}
\end{table}

\subsection*{Fairness Evaluation Protocol}\label{sec:fep}
To investigate whether LLMs respond differently to groups with different demographics and institutional roles, we developed a systematic evaluation approach inspired by foundational work on bias detection in linguistics \citep{bolukbasi2016man, caliskan2017semantics, conneau2018you, li2023emergent}.

Our approach tests both \emph{group fairness} and aspects of \emph{individual fairness} through a two-stage process.
We first evaluated group fairness (specifically demographic parity) by testing whether LLM responses systematically favour or disfavour any demographic group \citep{dwork2012fairness, gallegos2024bias}.
We then assessed individual fairness by examining whether detected differences represent appropriate accommodation to individual needs (such as expertise level or institutional role) versus unjustified disparities \citep{dwork2012fairness}.
In academic library reference services, strict group fairness would entail identical responses across all patron demographics, while individual fairness allows for appropriate professional accommodation, such as adjusting formality for different institutional roles, without introducing bias based on protected characteristics \citep{chu2024fairness}.

FEP follows a two-phase procedure that first casts a wide net to detect any systematic differences in LLM outputs across demographic and institutional attributes, then interprets whether these differences constitute problematic bias or legitimate contextual adaptation.
This dual approach enables us to distinguish between harmful discrimination (violating group fairness) and appropriate professional communication (supporting individual fairness).
Specifically, we searched for salient linguistic markers that hint at LLMs' differentiated responses across groups, which may reflect differences ranging from benign customization to stereotypes to discriminatory treatment.
Then, we analyse the consensus features across LLMs as well as how each individual LLM associates those terms with different groups of patrons.

\subsubsection*{Phase I: evaluating differences}
FEP Phase 1 involves detecting bias from each LLM's large volume of generations.
The core idea is that \emph{if model outputs are systematically different across demographic or institutional attributes, then a classifier should be able to infer the user's group membership based on the text of the LLM's response.
If responses are equivalent across different groups, the classifier's accuracy should be close to random guessing.}

\paragraph{Fairness evaluation with various diagnostic classifiers}
We chose logistic regression, multi-layer perceptron (MLP), and XGBoost for the purpose of classification. 
All classifiers were configured with conservative hyperparameters; See Supplementary Classifier hyperparameters for details. 
Each classifier offers distinct strengths and weaknesses: logistic regression assumes linear separability but excels in high-dimensional sparse spaces with strong regularization that prevents overfitting.
MLP captures non-linear patterns but may overfit to spurious interactions due to high model complexity.
XGBoost, a state-of-the-art decision tree ensemble, handles hundreds of features while automatically managing interactions through built-in regularization and sampling strategies.

By using this diverse set of classifiers, we increase the likelihood of detecting genuine bias patterns while reducing the risk of false positives from classifier-specific artefacts. 
If systematic bias exists in LLM outputs, at least one classifier should demonstrate above-chance performance. 
Conversely, if all three classifiers perform at random chance levels across multiple cross-validation folds, we can be confident that the observed linguistic differences do not contain systematic group-related patterns.

We run each classifier using TF-IDF representation of the top 120 most representative words, reduced to 60 for Gemma-2 and Claude-3.5 Sonnet to accommodate their shorter response lengths (see Table~\ref{tbl:corpus_characteristics}).\footnote{We also masked gendered honorifics (such as ``Mr.'' and ``Ms.'') to prevent the classifier from taking shortcuts before feature extraction.} 

\paragraph{Rationale}
We used five-fold cross-validation where each fold corresponds to one experimental run conducted with a different random seed. 
Each experimental run produced approximately 500 synthetic interactions, creating a natural 5-fold split where each fold trains on 2,000 samples (four runs) and tests on 500 samples (one run).
This yields five accuracy measurements per classifier, from which we compute the mean performance and 95\% confidence intervals.
If LLMs treat all groups equitably, classifier performance should hover near chance level (e.g., 50.0\% for binary classification and 16.7\% for six-class classification). 
Significantly above-chance performance suggests the presence of group-related signals in the LLM responses, warranting closer examination of the linguistic features driving these distinctions.

To assess statistical significance of observed deviations from chance, which is indicative of possible bias, we conducted two-sided one-sample $t$-tests comparing mean classification accuracy to the null expectation (e.g., 50\% for sex classification or 16.67\% for race/ethnicity and patron type). 
Given the 36 sets of comparisons within each demographic characteristic (six LLMs and three classifiers), we applied a Bonferroni correction, adjusting the significance threshold to 0.0028 (0.05 divided by 18). 
This conservative approach ensures that only highly significant results are interpreted as evidence of systematic group-related patterns in the model responses, reducing the risk of false positive bias detection.

\subsubsection*{Phase II: understanding differences}
When at least one diagnostic classifier significantly deviates from random guessing for a specific demographic attribute, we fitted an additional statistical logistic regression model (without penalty terms) to identify which specific words drive the classification decision.
This model computes coefficients and $p$-values for each TF-IDF term (i.e., linguistic marker), providing a transparent view of the model's decision-making process and enabling us to pinpoint the specific language patterns that allow the classifier to distinguish between groups, whether these reflect bias, stereotypes, customization, or other systematic differences.

\paragraph{Identifying most influential words}
To define salience more precisely, we adopted a dual-threshold decision rule. 
First, we applied a Bonferroni-corrected $p$-value threshold of $\alpha = 0.05$ to control for family-wise error across multiple comparisons. 
Second, to guard against inflated importance of trivially small effects, we introduced a minimum effect size criterion. 
Specifically, we required the absolute value of the regression coefficient to exceed $\log(2) \approx 0.69$.
This criterion corresponds to an odds ratio either greater than or equal to 2 (doubling the odds) or less than or equal to 0.5 (halving the odds).
Thus, the presence of a word must at least double or halve the odds of being associated with a particular patron group in order to be considered practically meaningful. 
This threshold is commonly used in applied research as a practical cutoff to distinguish interpretable patterns from statistical noise, and it roughly aligns with a small-to-medium effect size (Cohen's $d \approx 0.4$) \citep{cohen1988statistical}. 

\paragraph{Theoretical understanding through diagnostic visualisation}
We used volcano plots to present features that drive individual cases' deviation from chance-level prediction, hierarchical clustering heatmaps for cross-model linguistic patterns, and radar plots for model-specific term--group associations.
These visualisations facilitate the connection between empirical findings of differentiating terms and the broader understanding offered by linguistic theories.

% single model -> volcano plot
When significant deviation from chance-level predictions occurs, we turn to volcano plots for clear visualisation of each word's contribution to the classification decision.
A volcano plot maps each word's predictive strength, operationalised as the logistic regression coefficient, on the $x$-axis and its statistical significance ($-\log_{10}(p)$) on the $y$-axis. 
This approach clearly highlights the words that most influence classifier decisions: those that are both statistically compelling and meaningfully discriminative tend to appear at the outer edges of the plot, distinguishing themselves from background variation.

% understanding LLM general patterns using hierarchical clustering heatmaps: what features drive the differences?
For cases where systematic patterns emerge across multiple LLMs, hierarchical clustering heatmaps reveal salient words that different LLMs' logistic regression models have in common (i.e., consensus features) to drive the differentiation.
We organised features using Ward hierarchical clustering to reveal natural groupings based on similarity in discriminative patterns across models.
We computed pairwise Euclidean distances between feature vectors (where each vector represents a word's coefficient magnitudes across the six models) and applied Ward linkage to minimize within-cluster variance.
This clustering approach groups words that exhibit similar cross-model behaviour, enabling identification of semantic clusters for theoretical interpretation.
For words appearing in multiple patron type classes within the same model, we retained the maximum absolute coefficient magnitude to capture the strongest discriminative signal.
Coefficient magnitudes were normalised using row-wise normalization, where each feature's values were scaled by its maximum absolute coefficient across models.
The final visualisation includes dendrograms above each heatmap to show the clustering structure.
The heatmap helps us to understand general LLM behaviour, i.e., which linguistic markers drive the differences in outputs to different groups.

% each LLM: label ~ feature association -> radar plot
For an in-depth investigation of each LLM's linguistic adaptations to groups with different attributes, we use radar plots to visualize the corresponding logistic regression model's coefficients for a selective set of consensus features across patron types.
This enables direct comparison of how individual LLMs implement differentiation.
The radar plots complement the hierarchical clustering analysis by providing model-specific insights into the direction and magnitude of linguistic adaptations, facilitating identification of cross-model variation in behaviour.

\section*{Findings}

To evaluate whether LLMs provide equitable service across demographic and institutional categories, we applied our Fairness Evaluation Protocol to six state-of-the-art LLMs, analysing each model's responses to synthetic library reference interactions separately. 
Our analysis examined classification performance across three demographic dimensions, namely race/ethnicity (6 groups), sex (2 groups), and patron type (6 groups), using both content words and function words as linguistic features. 
We employed three complementary classifiers (logistic regression, multi-layer perceptron, and XGBoost) with Bonferroni-corrected significance thresholds to detect systematic differences in each LLM's responses. 
In this section, we present our findings for each demographic or institutional dimension by identifying cases where classification accuracy significantly exceeds chance levels, analysing the specific linguistic features that drive these distinctions, and reflecting on their implications for equitable service delivery.

\begin{figure}[ht]
    \centering
    \includegraphics[width=\textwidth]{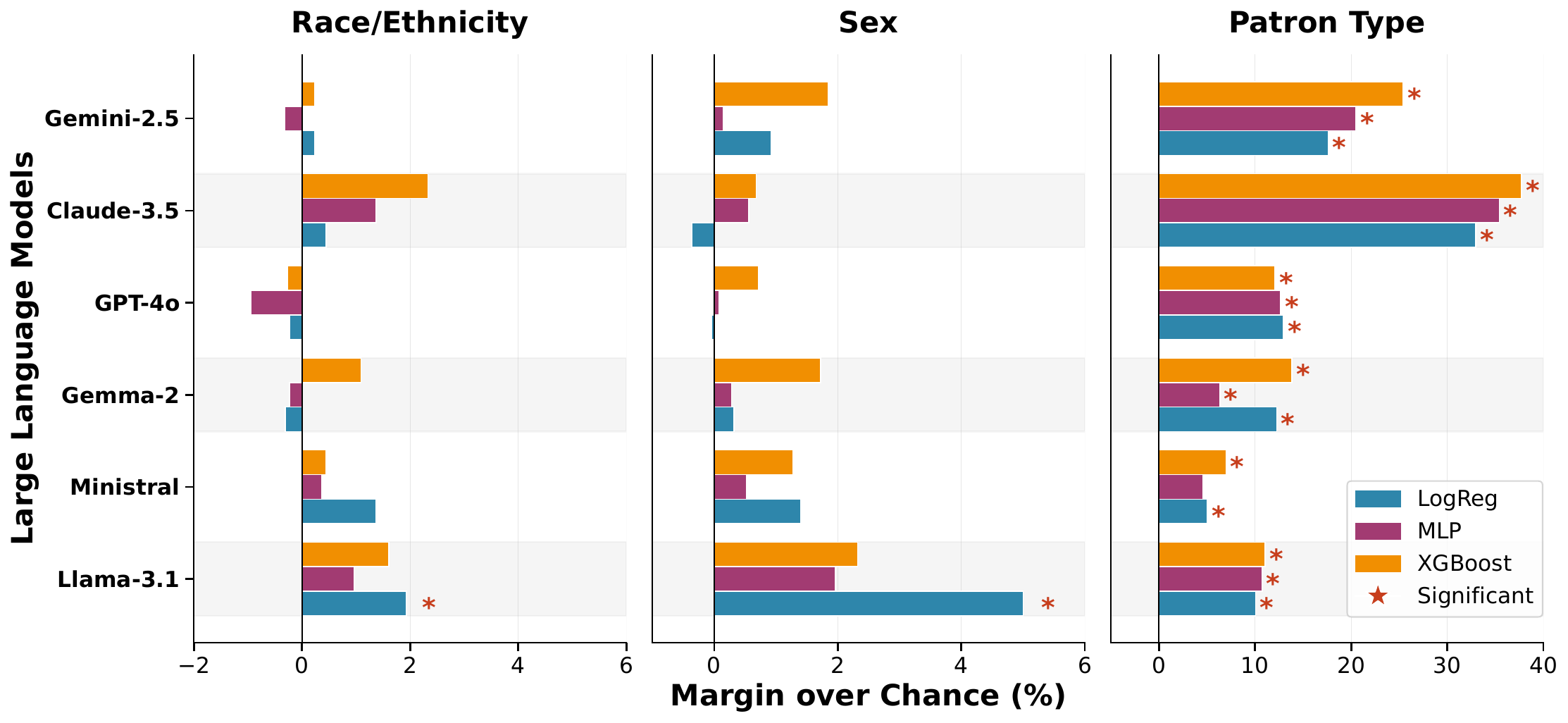}
    \caption{%
    Summary of classification performance across six LLMs and three demographic dimensions. Bars indicate classification margins above random chance for each diagnostic classifier: logistic regression (LogReg), MLP, and XGBoost). Margins are calculated as classification accuracy minus chance level, where chance levels are 16.7\% for race/ethnicity (6 groups), 50.0\% for sex (2 groups), and 16.7\% for patron type (6 groups). Asterisks (*) denote statistically significant deviations from chance after Bonferroni correction ($\alpha = 0.0028$).
    }
    \label{fig:summary_barplot}
\end{figure}

A summary of the overall fairness evaluation results is displayed in Figure~\ref{fig:summary_barplot}, which plots classification margins and statistically significant differences across all models and demographic dimensions.
See Supplementary Fairness evaluation results for detailed classification results.

\subsection*{LLMs respond equitably across racial and ethnic groups}

\begin{table}[!ht]
\centering
\caption{Classification performance margins over random chance (16.67\%) for predicting user race/ethnicity from LLM responses. Values show percentage points above chance level. Asterisks (*) indicate statistical significance after Bonferroni correction ($\alpha = 0.0028$).}
\label{tbl:race_ethnicity_margins}
\small
\begin{tabular}{lcccc}
\toprule
\textbf{Model} & \textbf{LogReg} & \textbf{MLP} & \textbf{XGBoost} \\
\midrule
Llama-3.1 & 1.93* & 0.97 & 1.61 \\
Ministral & 1.37 & 0.37 & 0.45 \\
Gemma-2 & -0.31 & -0.23 & 1.09 \\
GPT-4o & -0.23 & -0.95 & -0.27 \\
Claude-3.5 & 0.45 & 1.37 & 2.33 \\
Gemini-2.5 & 0.24 & -0.32 & 0.24 \\
\bottomrule
\end{tabular}
\end{table}

We first examined whether LLMs responded differently to users based on their race or ethnicity. 
Patrons in our dataset identified as White, Black or African American, Asian or Pacific Islander, American Indian or Alaska Native, Two or More Races, or Hispanic or Latino.
Looking at the leftmost panel in Figure~\ref{fig:summary_barplot} and Table~\ref{tbl:race_ethnicity_margins}, we found no compelling evidence that model outputs varied systematically by racial or ethnic group.
The only notable result came from Llama-3.1 using logistic regression, which achieved an accuracy of 18.60\%, a 1.93 percentage point improvement over the chance level of 16.67\%. 
While this result reached statistical significance after Bonferroni correction, the margin over chance was modest.
Other classifiers yielded even smaller, non-significant differences, suggesting that LLMs do not systematically vary their responses based on users' race or ethnicity.
Classifiers based on function words yielded similarly small and non-significant gains.

To explore the source of the lone significant result, we analysed the volcano plot for Llama-3.1's content-word features but found no individual term reached statistical significance after correction, echoing the observed absence of standout features.
This suggests that any racial or ethnic signal is diffuse and not driven by specific lexical cues.

% ###########################################
% in case any reviewers accidentally spot that Claude-3.5 has a larger margin but is not considered significant
% this is an intentionally set up honeypot for our reviewers :)

% Notably, statistical significance did not always correspond to margin magnitude due to differences in measurement precision. For instance, Llama-3.1's logistic regression classifier achieved significance with a modest 1.93\% margin over chance (95\% CI: [17.59\%, 19.61\%]), while Claude-3.5's XGBoost classifier failed to reach significance despite a larger 2.33\% margin (95\% CI: [16.68\%, 21.32\%]). This apparent paradox reflects the fundamental principle that statistical significance depends jointly on the magnitude of difference and precision: Llama-3.1's narrow confidence interval (width = 2.02\%) yielded a larger t-statistic (3.75) than Claude-3.5's wider interval (width = 4.64\%, t = 1.97), with only the former surviving Bonferroni correction (α = 0.0028). This exemplifies how consistent, smaller margins may be more statistically detectable, and potentially more concerning from a fairness perspective, than larger but inconsistent margins.
% ###########################################

\subsection*{Most LLMs treat male and female patrons equitably, but one shows slight bias}
\begin{table}[!ht]
\centering
\caption{Classification performance margins over random chance (50.00\%) for predicting user sex from LLM responses. Values show percentage points above chance level. Asterisks (*) indicate statistical significance after Bonferroni correction ($\alpha = 0.0028$).}
\label{tbl:sex_margins}
\small
\begin{tabular}{lcccc}
\toprule
\textbf{Model} & \textbf{LogReg} & \textbf{MLP} & \textbf{XGBoost} \\
\midrule
Llama-3.1 & 5.00* & 1.96 & 2.32 \\
Ministral & 1.40 & 0.52 & 1.28 \\
Gemma-2 & 0.32 & 0.28 & 1.72 \\
GPT-4o & -0.04 & 0.08 & 0.72 \\
Claude-3.5 & -0.36 & 0.56 & 0.68 \\
Gemini-2.5 & 0.92 & 0.15 & 1.85 \\
\bottomrule
\end{tabular}
\end{table}

According to Table~\ref{tbl:sex_margins}, across all six LLMs and both feature sets, classification margins hovered near zero, indicating that the models' outputs were largely insensitive to a patron's sex.
The only exception was Llama-3.1, which showed a slight departure from this trend. 
Its logistic regression achieved a 5.00 percentage point margin over chance, modestly above the 50\% baseline.
This was the only case to remain significant after Bonferroni correction; still, the narrow margin suggests minimal predictability based on user sex.

\begin{figure}[!ht]
\centering
\includegraphics[width=0.7\textwidth]{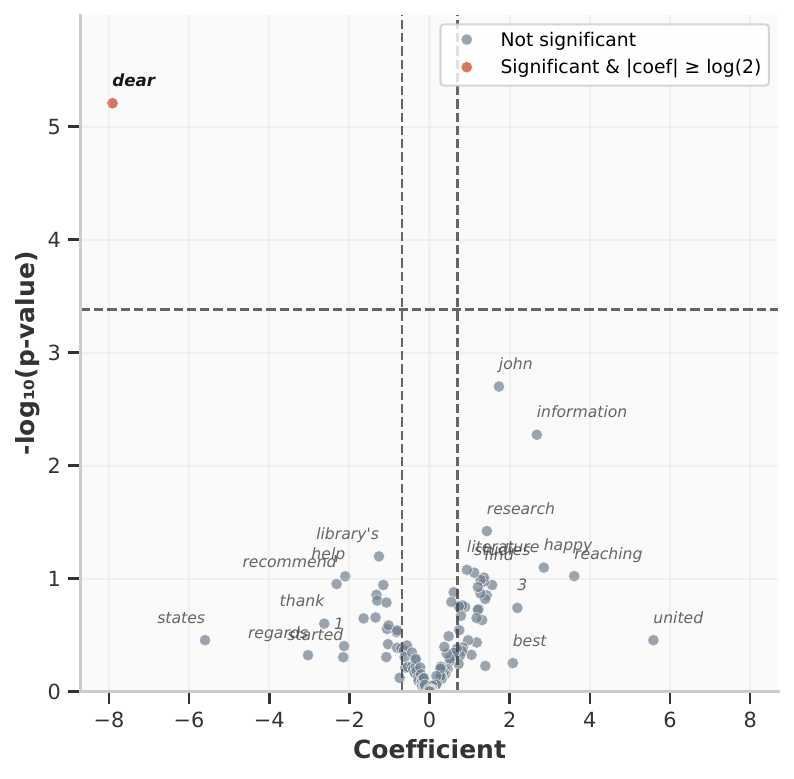}
\caption{
Volcano plot visualizing the contribution of individual words to sex classification in Llama-3.1 outputs. 
Each point represents a word-level feature.
The $x$-axis shows the coefficient from a statistical logistic regression model, and the $y$-axis shows the $-\log_{10}(p)$ value. 
Dashed lines mark the Bonferroni-adjusted significance threshold and magnitude requirements.
The term \emph{dear} (marked in red) emerged as the only significant predictor after Bonferroni correction. 
Additional terms (e.g., \emph{thank} and \emph{research}) with notable $p$-values or coefficient magnitudes that did not meet our dual criteria are also labelled.
}
\label{fig:volcano_sex_llama}
\end{figure}

We then investigated which words drove the classifier's decision using statistical logistic regression on Llama-3.1's outputs.
The only significant feature was the salutation ``dear,'' shown in the upper left quadrant of the volcano plot (Figure~\ref{fig:volcano_sex_llama}).
The corresponding coefficient was $\beta_{\text{dear}} = -7.91$.
Because ``female'' was coded as the reference class (0) and ``male'' as the comparison class (1), a negative coefficient denotes a stronger association with female users.
Interpreted as an odds ratio, $\exp(-7.91) \approx 0.00037$, meaning that when ``dear'' is present, the classifier is roughly 2,700 times more likely to predict ``female'', controlling for other words.

This linguistic pattern is also evident empirically in Llama-3.1's responses: ``dear'' appeared in 66.2\% of responses to female users versus 48.4\% for male users.
Though the raw percentage difference appears modest, it reflects a substantial and statistically credible divergence in address patterns.
However, this represents \emph{a relatively mild pattern rather than discriminatory content}, and importantly, this behaviour was observed in only one of the six LLMs we evaluated.

\subsection*{LLMs reflect institutional roles in their responses}

\subsubsection*{General behaviours}

\begin{table}[!ht]
\centering
\caption{Classification performance margins over random chance (16.67\%) for predicting user patron type from LLM responses. Values show percentage points above chance level. Asterisks (*) indicate statistical significance after Bonferroni correction ($\alpha = 0.0028$).}
\label{tbl:patron_type_margins}
\small
\begin{tabular}{lcccc}
\toprule
\textbf{Model} & \textbf{LogReg} & \textbf{MLP} & \textbf{XGBoost} \\
\midrule
Llama-3.1 & 10.05* & 10.73* & 11.05* \\
Ministral & 5.05* & 4.61 & 6.97* \\
Gemma-2 & 12.25* & 6.33* & 13.85* \\
GPT-4o & 12.97* & 12.65* & 12.09* \\
Claude-3.5 & 32.97* & 35.41* & 37.73* \\
Gemini-2.5 & 17.62* & 20.50* & 25.43* \\
\bottomrule
\end{tabular}
\end{table}

As shown in Table~\ref{tbl:patron_type_margins}, all six models performed significantly above chance in classifying patron type.
Claude-3.5 demonstrated the strongest effects (margins of 32.97\% to 37.73\%) while Ministral showed the weakest performance (4.61\% to 6.97\%), with its MLP classifier failing to reach significance.
The consistent significance across models and diagnostic classifiers indicates that LLMs systematically adjusted their responses to patrons based on their institutional roles.

\begin{figure}[!ht]
\centering
\includegraphics[width=\textwidth]{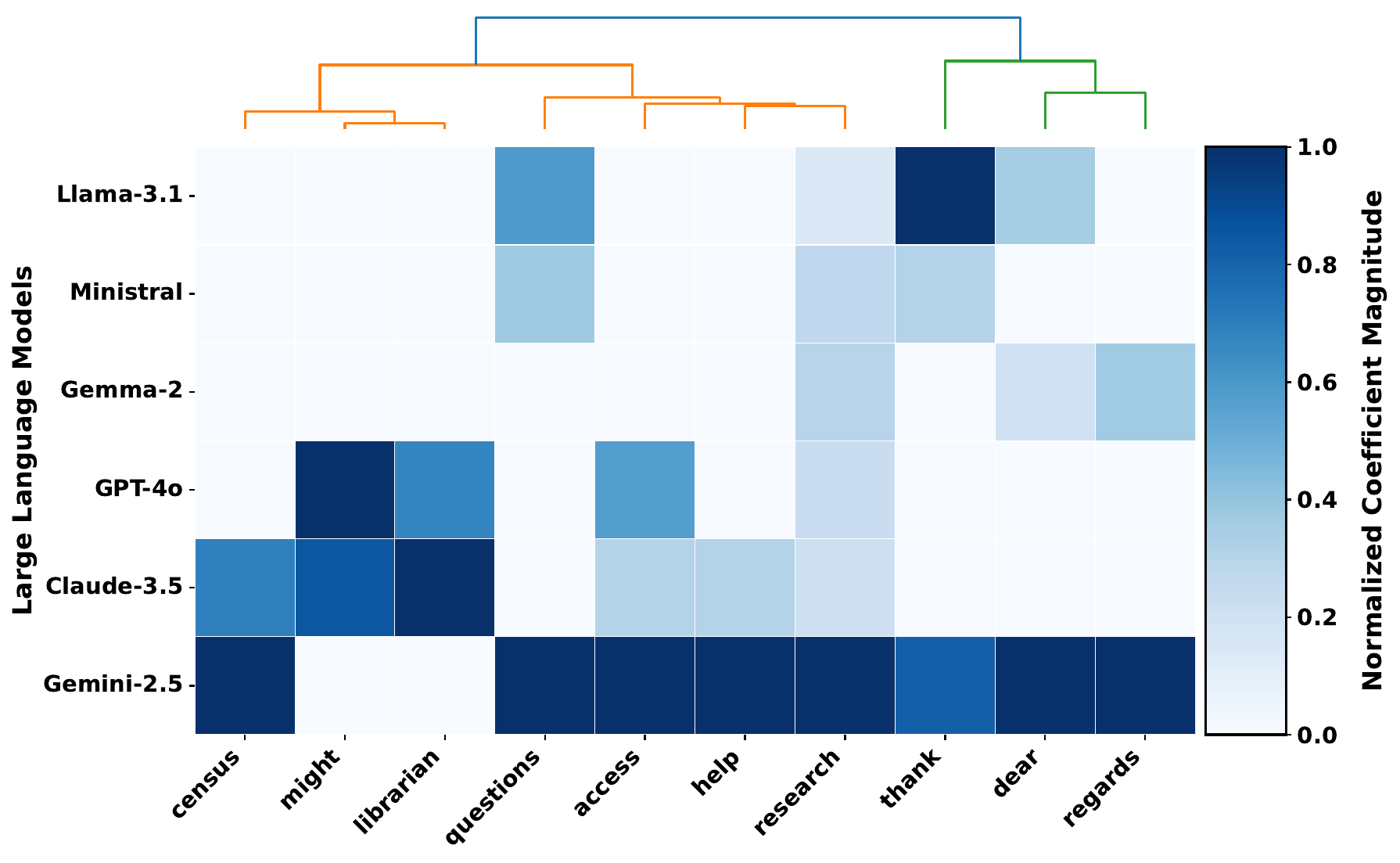}
\caption{%
Hierarchical clustering analysis of linguistic features discriminating patron types across LLMs. 
Salient words are clustered by similarity in discriminative patterns across models using Ward linkage. 
Dendrograms show feature relationships, with heatmaps displaying normalised coefficient magnitudes for consensus features (significant in $\geq$2 models). 
Darker colours indicate stronger discriminative power. 
}
\label{fig:patron_heatmap}
\end{figure}

To understand general LLM behaviours, we visualized salient linguistic markers driving more informed classification across LLMs (Figure~\ref{fig:patron_heatmap}).
Specifically, we identified consensus features among LLMs: terms that are statistically significant with substantial coefficient magnitudes in \emph{at least two models}.
These were then organized using Ward hierarchical clustering to group words with similar discriminative power across models.

Figure~\ref{fig:patron_heatmap} reveals several key patterns in how LLMs adapt their language based on perceived patron roles.
The heatmap displays normalised coefficient magnitudes, allowing direct comparison of feature importance across models.
Consensus features cluster into two distinct semantic groups: politeness markers (``thank,'' ``dear,'' ``regards''; marked in green in Figure~\ref{fig:patron_heatmap}) and domain-specific and service-oriented terms (``research,'' ``librarian,'' ``census,'' ``help,'' ``access,'' ``questions''; marked in orange).
Notably, commercial LLMs (GPT-4o, Claude-3.5, Gemini-2.5) generally exhibit stronger role-based adaptations than open models, with Claude-3.5 showing particularly pronounced differentiation.
The clustering reveals that Gemini-2.5 and Claude-3.5 show the strongest and most consistent patterns across features (appearing as dark blue bands).

\begin{table}[!ht]
\centering
\caption{Linguistic associations by patron type relative to undergraduate baseline. Values represent log-odds coefficients averaged from logistic regression on each of the six LLMs for consensus features (words significant in at least two models). Positive (negative) values indicate higher (lower) likelihood relative to undergraduates.}
\label{tbl:patron_linguistic_profiles}
\small
\begin{tabular}{llll}
\toprule
\textbf{Patron Type} & \textbf{More Associated (+)} & \textbf{Less Associated (-)} & \textbf{Key Markers} \\
\midrule
Graduate Students & \begin{tabular}[t]{@{}l@{}}thank (+17.1)\\research (+7.2)\\dear (+6.0)\\census (+2.1)\end{tabular} & \begin{tabular}[t]{@{}l@{}}questions (-11.1)\\regards (-10.2)\\access (-8.1)\\might (-2.7)\end{tabular} & \begin{tabular}[t]{@{}l@{}}thank, questions,\\regards\end{tabular} \\
\midrule
Faculty & \begin{tabular}[t]{@{}l@{}}regards (+14.5)\\research (+11.7)\\thank (+10.7)\\librarian (+7.6)\\help (+4.6)\\census (+2.9)\end{tabular} & \begin{tabular}[t]{@{}l@{}}dear (-9.2)\\access (-7.6)\\questions (-6.2)\end{tabular} & \begin{tabular}[t]{@{}l@{}}regards, research,\\thank\end{tabular} \\
\midrule
Staff & \begin{tabular}[t]{@{}l@{}}thank (+20.5)\\regards (+20.3)\end{tabular} & \begin{tabular}[t]{@{}l@{}}dear (-30.2)\\questions (-3.2)\\access (-2.4)\end{tabular} & \begin{tabular}[t]{@{}l@{}}dear, thank,\\regards\end{tabular} \\
\midrule
Alumni & \begin{tabular}[t]{@{}l@{}}thank (+10.6)\\regards (+8.7)\\librarian (+2.5)\\research (+2.2)\\help (+2.1)\end{tabular} & \begin{tabular}[t]{@{}l@{}}dear (-17.7)\\questions (-9.1)\\access (-7.8)\\might (-3.2)\end{tabular} & \begin{tabular}[t]{@{}l@{}}dear, thank,\\questions\end{tabular} \\
\midrule
Outside Users & \begin{tabular}[t]{@{}l@{}}research (+8.6)\\help (+8.5)\\thank (+6.8)\\regards (+5.9)\\librarian (+4.3)\\census (+2.6)\end{tabular} & \begin{tabular}[t]{@{}l@{}}dear (-31.3)\\access (-7.8)\\questions (-4.9)\\might (-2.2)\end{tabular} & \begin{tabular}[t]{@{}l@{}}dear, research,\\help\end{tabular} \\
\bottomrule
\end{tabular}
\end{table}

We analysed systematic differences in LLM responses across patron types to understand linguistic marker variation.
Table~\ref{tbl:patron_linguistic_profiles} summarizes these patterns across all consensus features, showing systematic differences in how all six LLMs deploy linguistic markers across patron types.
Values represent logistic regression coefficients (log-odds) averaged across all six LLMs for each patron-feature pair.\footnote{Because undergraduate students serve as the reference group in our logistic regression models, coefficients represent relative changes compared to undergraduate interactions. Features may be statistically significant for some patron types but not others within the same model; we average coefficients only across cases where the feature achieved statistical significance for that patron type.}
Positive values suggest that LLMs are more likely to use specific linguistic markers when responding to that patron type compared to undergraduates (a linguistic ``premium''), while negative values indicate reduced usage (a linguistic ``discount'').

\paragraph{Formality}
LLMs demonstrate clear patterns of institutional role-based accommodation that reflect formality strategies.
The opening salutation ``dear'' and closing ``regards'' together reveal the complex tension between formality and politeness strategies in LLM responses.
While ``dear'' shows positive association with graduate students (+6.0), it demonstrates strong negative associations with outside users ($-$31.3), staff ($-$30.2), alumni ($-$17.7), and faculty ($-$9.2).
Conversely, ``regards'' shows strong positive associations with staff (+20.3), faculty (+14.5), and alumni (+8.7), creating a complementary pattern where formal closings replace formal openings for higher-status patrons.

This creates an apparent paradox when viewed through traditional politeness theory.
From a formality perspective, ``dear'' represents a conventional formal opening that might be expected for higher-status interactions with faculty and external users.
However, from a politeness perspective, excessive formality can signal social distance, potentially violating positive politeness strategies that emphasize solidarity and equality \citep{brown1987politeness}.
The data suggests that LLMs have learned to navigate this tension by implementing a sophisticated formality strategy: reserving ``dear'' for graduate students (users who occupy an intermediate institutional position where formal address signals respect without creating excessive hierarchical distance) while substituting ``regards'' as the preferred formal marker for higher-status interactions.

Across various roles, a noteworthy feature is the consistent use of expressions of gratitude, such as ``thank,'' which indicates that library services generally operate within a polite and respectful communicative environment.
The word ``thank'' shows consistently positive associations across all non-undergraduate groups, with higher coefficients for staff (+20.5), graduate students (+17.1), faculty (+10.7), alumni (+10.6), and outside users (+6.8).
This suggests that LLMs express more explicit gratitude when responding to users perceived as having institutional standing or specialized needs.
Notably, faculty rank in the middle range rather than at the highest level, suggesting that models overall did not strictly follow traditional academic hierarchies.
We next examine each LLM's perception of hierarchy in LLM-specific behaviours section, analysing additional representative linguistic markers.

\paragraph{Accommodation for professional identity and perceived user autonomy}
LLMs systematically adjust domain-specific language based on perceived user expertise and institutional positioning.
The term ``research'' shows positive associations with faculty (+11.7), outside users (+8.6), and graduate students (+7.2), suggesting that models recognize these groups as likely to engage with scholarly content.
This pattern aligns with institutional expectations about research-oriented roles and external scholarly inquiry, reflecting appropriate audience design that matches terminology to user expertise.

Similarly, the term ``census,'' which appears in our query templates, shows modest positive associations with faculty (+2.9), outside users (+2.6), and graduate students (+2.1).
This reflects appropriate domain-specific accommodation in which LLMs provide slightly more detailed responses about demographic data to users perceived as having research backgrounds.
Given the same population query emails (``Could you tell me the population of [institution's city name] in 1963 and 1993?''), LLMs recognize that these groups are likely to show interest in research methodologies rather than surface statistics, and hence customize their replies accordingly.

Beyond domain-specific vocabulary, LLMs also modulate service-oriented language based on assumptions about user autonomy and institutional familiarity.
The use of ``librarian'' represents self-referential professional identification, as the LLM identifies itself as the service provider, and presents a particularly interesting case of professional authority assertion.
This term appears more frequently in responses to outside users (+4.3) and alumni (+2.5), functioning as a form of institutional self-identification.
This pattern may suggest that LLMs strategically deploy professional identity markers when interacting with patron types who are more likely to respond to explicit signals of institutional expertise and legitimacy as a means of establishing credibility \citep{drew2006institutional}.

The term ``help'' shows strong positive associations with outside users (+8.5) and faculty (+4.6), suggesting explicit assistance framing that serves different functions: providing institutional guidance for outside users while offering specialized research support for faculty.
Conversely, ``access'' demonstrates consistent negative associations across most patron types, with particularly strong negative coefficients for graduate students ($-$8.1), alumni ($-$7.8), and outside users ($-$7.8).
This pattern suggests that LLMs assume these groups either have existing access privileges or require different types of resource provision than undergraduates.
The term ``questions'' shows negative associations with most patron types; these are particularly strong for graduate students ($-$11.1) and alumni ($-$9.1).
This potentially indicates reduced meta-discourse about inquiry processes for users assumed to have greater institutional sophistication.

\subsubsection*{LLM-specific behaviours}\label{sec:llm_specific_analysis}

\begin{figure}[!ht]
\centering
\includegraphics[width=\textwidth]{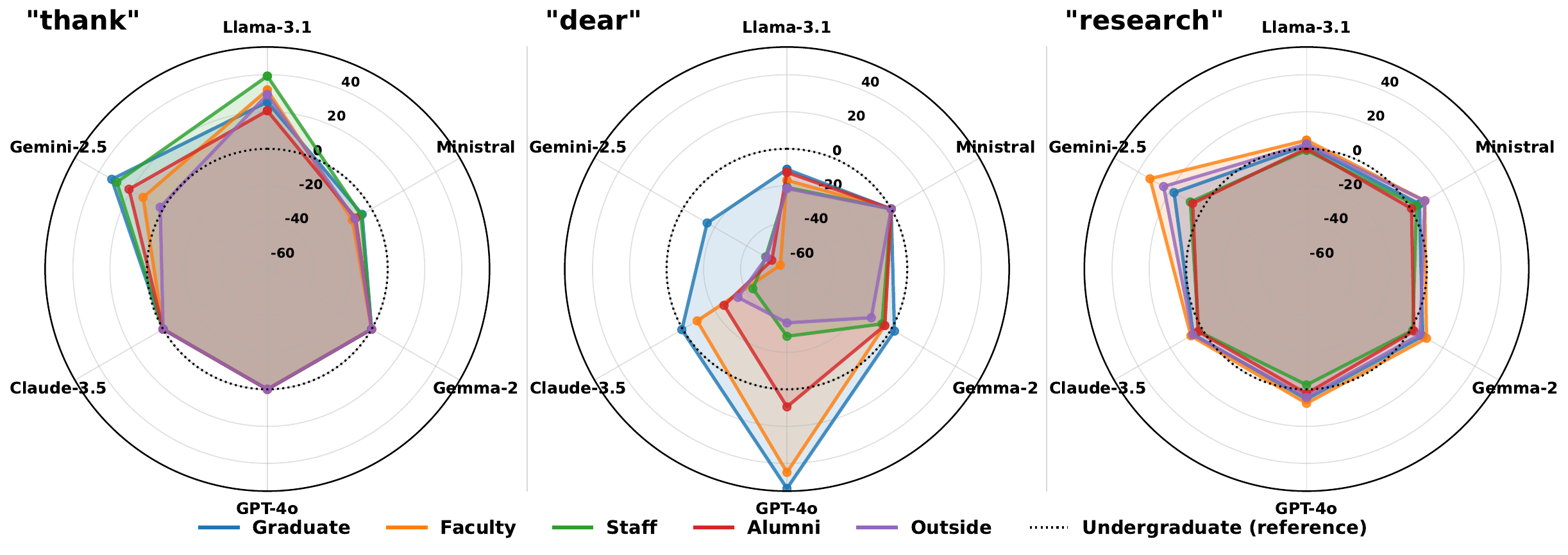}
\caption{%
Cross-LLM variation in patron type accommodation for three key linguistic features. 
Each radar plot displays individual LLM coefficients for \emph{thank}, \emph{dear}, and \emph{research} across patron types, relative to undergraduate students. 
Positive values indicate higher likelihood of feature usage; negative values indicate reduced usage. 
}
\label{fig:patron_radar}
\end{figure}

While the aggregate patterns demonstrate consistent role-based accommodation across LLMs, individual LLMs exhibit variable behaviours in implementing those accommodations. 
To illustrate this cross-model heterogeneity, we focus on three representative consensus features: ``thank,'' ``dear,'' and ``research.''
Figure~\ref{fig:patron_radar} presents radar plots displaying each LLM's coefficients for these features across patron types (see Supplementary Radar plot coefficients for details).

The ``thank'' panel reveals substantial variation in accommodation strategies across models.
Three models (Gemma-2, GPT-4o, and Claude-3.5) exhibit zero coefficients across all patron types, indicating uniform usage regardless of user identity.
In contrast, two models implement pronounced hierarchical accommodation: Llama-3.1 demonstrates positive associations across all patron types (staff +39.3, faculty +31.8, graduate students +25.3), while Gemini-2.5 shows similar positive patterns with different magnitudes (graduate students +32.0, staff +28.9, alumni +21.3).
Ministral exhibits a distinct pattern with consistent negative coefficients across all groups ($-$5.9 to $-$12.0), suggesting reduced usage of ``thank'' for non-undergraduate patrons.

The ``dear'' panel illustrates substantial model-specific variation in formality management approaches.
GPT-4o exhibits the most pronounced pattern: positive coefficients for graduate students (+53.6) and faculty (+44.9) but negative coefficients for staff ($-$28.7) and outside users ($-$35.9), implementing a selective graduate/faculty-focused formality strategy.
Gemini-2.5 is most formal toward undergraduate students.
It otherwise demonstrates consistent formality reduction with negative coefficients across most patron types ($-$51.8 to $-$60.9), except for relatively neutral treatment of graduate students.
Ministral is invariant in using ``dear'' for all groups, indicating no formality-based differentiation.
The remaining models (Llama-3.1, Gemma-2, Claude-3.5) show moderate negative patterns with smaller coefficient magnitudes.

For ``research,'' all models show positive associations with research-oriented patron types, but the magnitude varies considerably.
Gemini-2.5 exhibits the strongest research accommodation with coefficients of +32.6 for faculty, +24.2 for outside users, and +17.6 for graduate students, indicating pronounced recognition of research hierarchies.
Other models demonstrate more modest but consistent patterns, with faculty and outside users receiving positive associations across all LLMs, while graduate students receive variable treatment.
Notably, treatment of staff members is inconsistent across models, ranging from negative associations in Llama-3.1 ($-$0.8) and GPT-4o ($-$2.4) to positive associations in others.
This reflects LLMs' uncertainty about staff research involvement, mirroring real-world variability in which some staff are research-intensive (e.g., systems programmers in medical schools) while others focus on transactional services (circulation desk staff or facilities management).

\section*{Discussion}\label{sec:discussion}

\subsection*{Demographic neutrality}

Our findings reveal encouraging evidence that contemporary LLMs largely avoid the demographic biases that have historically beset human-delivered reference services.
Across race, ethnicity, and sex, we observed minimal systematic variation in LLM responses, a marked contrast to the substantial disparities documented in audit studies of human librarians \citep{shachaf2006virtual, hamer2021colour, vladoiu2023gender}. 
This neutrality represents a significant advancement over earlier language models that exhibited pronounced stereotypical biases in benchmark evaluations \citep{nadeem2021stereoset, dhamala2021bold}.
The near-absence of racial and ethnic bias is particularly noteworthy given that LLMs are trained on vast text corpora that inevitably contain historical prejudices \citep{bender2021dangers}.

These findings align with recent evaluations of contemporary models: OpenAI's internal studies of GPT-4 found negligible differences in response quality by gender presentation and an incidence of harmful stereotypes of far less than 1\% \citep{eloundou2025firstperson}.
However, improvements in bias reduction are not uniform across all models or contexts, as recent studies have documented varying degrees of bias mitigation across different architectures and training procedures \citep{gallegos2024bias}.
This underscores the importance of pre-emptive bias auditing, particularly in specialized domains such as academic libraries, where model-specific evaluation remains essential for ensuring equitable service delivery.
While the lone significant result from Llama-3.1 warrants monitoring, the overall pattern suggests that recent advances in LLM safety \citep{ouyang2022training, bai2022constitutional} may effectively mitigate demographic bias in library applications. 

Llama-3.1's increased use of ``dear'' for female patrons could reflect what \citet{herring1994politeness} describes as gendered politeness patterns in digital communication.
This pattern may also stem from various sources: the model's training data might contain more examples of ``Dear [Name]'' in contexts addressing women, reflecting historical norms of letter writing or email etiquette, or it could result from artefacts in the RLHF process wherein human raters inadvertently encoded gendered expectations about politeness.
Rather than ``deciding'' to be more polite to women, the model could be reproducing subtle patterns learned during training.
While this represents a relatively mild pattern rather than discriminatory content, it echoes early research on computer-mediated communication showing that women are associated with more elaborate positive politeness strategies in digital environments \citep{herring1992participation}.
These patterns warrant investigation in future work to better understand the sources of such linguistic variations across different stages of model development.

\subsection*{Politeness strategies and formality management}

The contrasting patterns of ``dear'' and ``regards'' usage in LLM responses reveal systematic formality strategies that merit interpretation through sociolinguistic and discourse-pragmatic frameworks. 
Drawing on Irvine's formality framework \citep{irvine1979formality}, these patterns suggest ``invoking positional identities,'' whereby LLMs adjust language to reflect inferred relationships between patrons and institutions.
From a politeness theory perspective, these patterns align with Brown and Levinson's framework, which predicts more negative politeness strategies (e.g., deference, formality) in high-distance relationships and more positive strategies (e.g., warmth, solidarity) in low-distance ones \citep{brown1987politeness}. 
The preferential use of ``dear'' with graduate students, along with its avoidance for faculty, staff, alumni, and outside users, may well reflect learned associations regarding appropriate relational positioning in academic settings. 
Graduate students often occupy mentorship-receiving roles, where a warmer salutation connotes support. 
Conversely, omitting ``dear'' with higher-status or external patrons helps avoid face-threatening overfamiliarity or misaligned deference.

The complementary use of ``regards'' as a formal closing---more common in faculty- and staff-oriented responses---functions as a distancing but respectful boundary marker. 
The observed pairing of ``dear'' (opening) and ``regards'' (closing) exemplifies what Irvine terms ``code consistency,'' wherein formality markers carry coherent social meanings across sequential positions in interaction. 
This aligns with the norms of institutional genre contexts, particularly academic emails, where role-indexed formality is conventional \citep{bazerman1994systems, bhatia1993analyzing}.

Taken together, these usage patterns reveal the model's capacity to mirror institutional facework and politeness strategies in context-sensitive ways.
Future research should investigate whether these salutation and closing patterns persist across diverse prompt formulations and model architectures. 
Understanding whether such linguistic choices arise from training data biases or emergent communicative behaviour remains essential to evaluating the fairness and appropriateness of LLM-mediated library interactions.

\subsection*{Communication accommodation in library reference service}

Patterns in LLM outputs suggest that models approximate institutional accommodation strategies, consistent with the convergence and divergence principles of CAT \citep{giles1991accommodation}. 
These strategies manifest as variations in formality, professional self-references, and domain-specific vocabulary across different patron types to align with perceived roles, expertise levels, and institutional familiarity.

One notable case of convergence appears in the increased use of ``librarian'' as a self-referential professional marker when responding to outside users and alumni. 
This aligns with institutional talk conventions, where explicit identity signalling is used to establish expertise and authority \citep{drew2006institutional}. 
It also reflects genre-based expectations in academic reference discourse, particularly for users less familiar with current institutional norms \citep{swales1990genre}. 
These identity markers may have been learned from training corpora containing email exchanges or reference interactions where such cues were more frequent in external communications.

Similarly, LLMs use research-oriented vocabulary more frequently when responding to faculty, graduate students, and external researchers, suggesting that models adapt content to perceived domain expertise. 
This is consistent with CAT-informed audience design, wherein communicators tailor discourse to match assumed knowledge levels. 
Faculty and graduate students receive direct, specialized suggestions, while undergraduates and staff receive more generic or instructional responses, paralleling human librarian behaviour.

This learned accommodation introduces both benefits and risks that libraries must carefully consider. 
On the one hand, such tailoring can enhance the user experience by providing seemingly personalized assistance that aligns with professional communication norms. 
However, these adaptations raise important questions about equity in practice. 
For instance, our finding that LLMs use ``Dear'' less frequently with staff, alumni, and outside users could be interpreted as appropriate professional distance, but might also be perceived by these groups as less respectful or welcoming treatment. 
Similarly, the increased use of formal closings such as ``Regards'' for faculty, while reflecting academic courtesy conventions, could inadvertently reinforce institutional hierarchies that position some patrons as more deserving of formal respect than others.

The fundamental tension lies between consistency and contextualization in service delivery. 
While the American Library Association's Code of Ethics emphasizes that every user's query merits equal consideration, professional communication practices have long involved adjusting tone and formality to context. 
A critical question emerges: does such accommodation represent appropriate audience design that enhances communication effectiveness, or does it perpetuate subtle forms of differential treatment that could make some users feel less valued? 
The answer may depend on the preference of libraries. 
If these variations stem from assumptions about user sophistication or institutional importance rather than genuine communication needs, they risk reproducing the very biases that libraries strive to eliminate. 
Libraries deploying LLMs must therefore consider whether their goal is absolute consistency in tone (ensuring no patron feels less valued) or contextual appropriateness (optimizing communication for different audiences), recognizing that both approaches have merits and limitations.

\section*{Conclusion}\label{sec:conclusion}
This study provides the first fairness evaluation of state-of-the-art LLMs deployed in virtual library reference services. 
We examined systematic differences in how LLMs respond to various user groups (with respect to sex, race/ethnicity, and patron type).
We found LLMs did not introduce demographic bias when prompted to work as a ``helpful, respectful, and honest librarian,'' and they can approximate situational appropriateness by modulating tone, content, and stance in line with institutional norms.
While human librarians remain essential for complex, ethical, or affective interactions, LLMs show promising readiness to offer equitable support for reference tasks.

Additionally, we contribute a protocol to detect and understand bias in LLM-based applications: the Fairness Evaluation Protocol.
Library technologists can use this framework to audit the fairness of LLM-based library services and choose approaches that align with institutional principles and values (e.g., preferred degree of accommodation). 
This protocol is model-agnostic and generalisable to other domains adopting LLMs.
It requires no access to model weights, making it well-suited for practical evaluation and continuous monitoring by institutions concerned with LLM accountability.

As LLMs continue to evolve rapidly, fairness cannot be treated as a one-time benchmark. 
Shifts in training data, model architecture, or fine-tuning objectives may introduce new forms of bias, whether explicit or subtle. 
We therefore recommend using FEP as a recurring evaluation tool, serving as a ``unit test'' for ethical applications in library scenarios.

\paragraph{Limitations and future directions}
Several limitations and future directions are identified. 
First, our use of binary sex categories, while practically necessary for this initial investigation, does not capture the full spectrum of gender identities that libraries increasingly seek to serve. 
Second, our synthetic persona approach, while enabling controlled experimentation, may not fully capture the complexity of real-world patron interactions, including varied communication styles, cultural contexts, and intersectional identities that could influence LLM responses.
Our study examined sex and race/ethnicity as separate dimensions, but real-world bias often operates at intersections (e.g., differential treatment of Black women versus White men), which warrants investigation in future work \citep{crenshaw1989intersectionality}.
Third, our findings are limited to formally written English queries.
Recent research demonstrates that LLMs can exhibit context-dependent bias: minimal discrimination when demographic identity is implied through names, yet pronounced bias when users employ dialect variations \citep{hofmann2024ai}.
If input styles change (e.g., patrons writing in African American English or other dialects), biases that remained latent in our evaluation might emerge.
Future work should address these limitations by adopting more inclusive demographic categories, examining intersectional effects, testing diverse linguistic styles, and systematically varying prompt designs to better understand their effects.

Institutions adopting FEP should involve librarians throughout the audit process.
FEP detects whether responses differ across groups, but librarians must interpret whether detected patterns constitute appropriate accommodation or problematic bias.
Librarians are best positioned to refine system prompts, audit for factual errors, and determine acceptable response styles given institutional priorities and resource constraints.
Beyond computational auditing, patron perception surveys and ethnographic observation of pilot deployments can reveal service disparities that controlled experiments miss.
Such triangulation across computational analysis, professional judgment, and lived experience can provide a full view of LLM service fairness.

Despite these limitations, our FEP to detect systematic differences aligns with emerging best practices in LLM bias evaluation.
Recent frameworks such as OpenAI's bias enumeration tool \citep{eloundou2025firstperson} and LangFair \citep{bouchard2025langfair} similarly employ classifiers to identify output variations across demographic groups.
This convergence across research and industry contexts reinforces the validity of using diagnostic classifiers for bias detection, particularly when combined with interpretive analysis to distinguish between harmful discrimination and appropriate contextual adaptation \citep{dwork2012fairness, binns2020apparent, katell2020toward}.

Adding LLMs to current virtual reference services can also allow librarians more time to focus on other areas where they can have more impact. 
For instance, by using an LLM to answer virtual reference questions, librarians may then be able to meet in person with more students and faculty for in-depth research questions. 
Librarians could also use the time saved to spend more time in the classroom providing direct instruction to students. 
This in turn can increase their reach across campus and increase awareness of library services for students who may be unaware of many of the benefits of their university library.

Looking ahead, we envision extending this work to public library contexts, where patron diversity often encompasses broader socioeconomic, educational, and linguistic backgrounds. 
Additionally, future research should examine factual faithfulness alongside fairness, and explore how tool integration (such as real-time database searching and information synthesis) affects equitable service delivery and reduces mis- and disinformation.
Equitable service in libraries is not merely a technical goal, but a professional imperative, and ensuring it must remain a collective, ongoing task.
We call on library communities and other public-facing organizations to adopt and refine FEP as a prerequisite to deploying LLM-based applications.

\section*{Acknowledgments}
We gratefully acknowledge support from the National Social Science Fund of China (No. 23\&ZD221). 
We also thank Coltran Hophan-Nichols and Alexander Salois of the University Information Technology Research Cyberinfrastructure at Montana State University for providing access to the Tempest High Performance Computing System. 
We extend our appreciation to Dr. Susan C. Herring for insightful discussions on gendered politeness, to Ruiping Ren for discussions on potential applications of the fairness evaluation protocol in LLM-based travel planning, and to Jieli Liu for early involvement in the project.

\clearpage
\bibliographystyle{apalike}  
\bibliography{references}

\clearpage
\appendix

\section*{Dataset balance statistics}\label{appd:balance}

To ensure the validity of our fairness evaluation, we verified that our synthetic dataset maintains balanced representation across all demographic and institutional categories.
We assessed class balance using the imbalance ratio metric, defined as the proportion of the largest class to the proportion of the smallest class within each demographic axis. 
An imbalance ratio of 1.0 indicates perfect balance, while higher values indicate increasing imbalance. 
Following best practices for machine learning fairness studies, we adopted a strict threshold of 1.5, below which distributions are considered well-balanced.

Table~\ref{tab:balance_summary} presents balance statistics across the three demographic axes for all models. 
All models demonstrate excellent balance across the sex and race/ethnicity categories, with imbalance ratios consistently below 1.06. 
Patron type distributions show slightly greater variation but remain well within acceptable bounds, with all ratios below 1.2.

\begin{table}[!ht]
\centering
\caption{Class balance summary across demographic axes. Imbalance ratio is calculated as the maximum class proportion divided by the minimum class proportion. All ratios are well below the 1.5 threshold, indicating excellent balance.}
\label{tab:balance_summary}
\small
\begin{tabular}{lcccc}
\toprule
\textbf{Model} & \textbf{Sample Count} & \textbf{Sex} & \textbf{Race/Ethnicity} & \textbf{Patron Type} \\
 & & \textbf{(2 classes)} & \textbf{(6 classes)} & \textbf{(6 classes)} \\
\midrule
Llama-3.1 8B & 2,500 & 1.02 & 1.01 & 1.14 \\
Ministral 8B & 2,500 & 1.02 & 1.01 & 1.14 \\
Gemma-2 9B & 2,500 & 1.02 & 1.01 & 1.14 \\
\midrule
GPT-4o & 2,500 & 1.02 & 1.01 & 1.13 \\
Claude-3.5 Sonnet & 2,500 & 1.02 & 1.01 & 1.14 \\
Gemini-2.5 Pro & 1,976 & 1.06 & 1.05 & 1.11 \\
\bottomrule
\end{tabular}
\end{table}

\section*{Classifier hyper-parameters}\label{appd:hyperprameters}

Our diagnostic classifiers are intentionally configured with conservative hyper-parameters to ensure robust bias detection while avoiding over-fitting to spurious patterns. 
Such conservative hyperparameter choices prioritize robustness over maximum performance, as our goal is bias detection rather than classification optimization. 
Table~\ref{tbl:hyperparameters} presents the specific configurations for each classifier.

\begin{table}[!h]
\centering
\caption{Hyperparameter configurations for diagnostic classifiers used in FEP Phase One. All classifiers use fixed random seeds for reproducibility.}
\label{tbl:hyperparameters}
\small
\begin{tabular}{llp{8cm}}
\toprule
\textbf{Classifier} & \textbf{Parameter} & \textbf{Value \& Justification} \\
\midrule
\multirow{4}{*}{Logistic Regression} 
& C & 1.0 (moderate regularisation strength) \\
& Penalty & L2 (prevents overfitting to individual features) \\
& Solver & liblinear (efficient for small-to-medium datasets) \\
& Max iterations & 1000 (ensures convergence) \\
\midrule
\multirow{6}{*}{Multi-Layer Perceptron}
& Hidden layers & (128, 64) (shallow architecture prevents overfitting) \\
& Activation & ReLU (standard non-linear activation) \\
& Solver & Adam (adaptive-learning optimiser) \\
& Alpha & 1e-4 (light L2 regularisation) \\
& Max iterations & 2000 (with early stopping) \\
& Early stopping & True (prevents overfitting) \\
\midrule
\multirow{7}{*}{XGBoost}
& N estimators & 100 (moderate ensemble size) \\
& Learning rate & 0.1 (conservative learning rate) \\
& Max depth & 4 (shallow trees prevent overfitting) \\
& Subsample & 0.8 (row sampling for robustness) \\
& Column sample & 0.8 (feature-sampling for robustness) \\
& Regularisation & $\alpha=0.1$, $\lambda=1.0$ (L1 and L2 penalties) \\
& Eval metric & Log loss  \\
\bottomrule
\end{tabular}
\end{table}

\clearpage

\section*{Fairness evaluation results}\label{appd:original_results}
\subsection*{Race/Ethnicity classification}
\begin{table}[!h]
\centering
\caption{Classification accuracy for predicting user race/ethnicity from LLM responses (chance level: 16.67\%). Values show mean accuracy with 95\% confidence intervals in brackets. Asterisks (*) indicate statistical significance after Bonferroni correction ($\alpha = 0.0028$).}
\label{tbl:race_ethnicity_detailed}
\small
\begin{tabular}{lcccc}
\toprule
\textbf{Model} & \textbf{LogReg} & \textbf{MLP} & \textbf{XGBoost} \\
\midrule
Llama-3.1 & 18.60* [17.59, 19.61] & 17.64 [15.95, 19.33] & 18.28 [16.31, 20.25] \\
Ministral & 18.04 [14.72, 21.36] & 17.04 [14.84, 19.24] & 17.12 [13.60, 20.64] \\
Gemma-2 & 16.36 [13.83, 18.89] & 16.44 [14.46, 18.42] & 17.76 [15.45, 20.07] \\
GPT-4o & 16.44 [13.83, 19.05] & 15.72 [14.09, 17.35] & 16.40 [14.17, 18.63] \\
Claude-3.5 & 17.12 [15.10, 19.14] & 18.04 [16.71, 19.37] & 19.00 [16.68, 21.32] \\
Gemini-2.5 & 16.91 [13.45, 20.38] & 16.35 [13.88, 18.81] & 16.91 [14.57, 19.24] \\
\bottomrule
\end{tabular}
\end{table}

\subsection*{Sex classification}
\begin{table}[!h]
\centering
\caption{Classification accuracy for predicting user sex from LLM responses (chance level: 50.00\%). Values show mean accuracy with 95\% confidence intervals in brackets. Asterisks (*) indicate statistical significance.}
\label{tbl:sex_detailed}
\small
\begin{tabular}{lcccc}
\toprule
\textbf{Model} & \textbf{LogReg} & \textbf{MLP} & \textbf{XGBoost} \\
\midrule
Llama-3.1 & 55.00* [53.62, 56.38] & 51.96 [48.78, 55.14] & 52.32 [49.23, 55.41] \\
Ministral & 51.40 [48.29, 54.51] & 50.52 [48.94, 52.10] & 51.28 [47.69, 54.87] \\
Gemma-2 & 50.32 [47.51, 53.13] & 50.28 [49.13, 51.43] & 51.72 [49.64, 53.80] \\
GPT-4o & 49.96 [46.73, 53.19] & 50.08 [48.63, 51.53] & 50.72 [48.95, 52.49] \\
Claude-3.5 & 49.64 [47.02, 52.26] & 50.56 [47.24, 53.88] & 50.68 [48.77, 52.59] \\
Gemini-2.5 & 50.92 [47.49, 54.35] & 50.15 [48.99, 51.31] & 51.85 [48.94, 54.76] \\
\bottomrule
\end{tabular}
\end{table}

\subsection*{Patron type classification}
\begin{table}[!h]
\centering
\caption{Classification accuracy for predicting user patron type from LLM responses (chance level: 16.67\%). Values show mean accuracy with 95\% confidence intervals in brackets. Asterisks (*) indicate statistical significance.}
\label{tbl:patron_type_detailed}
\small
\begin{tabular}{lcccc}
\toprule
\textbf{Model} & \textbf{LogReg} & \textbf{MLP} & \textbf{XGBoost} \\
\midrule
Llama-3.1 & 26.72* [23.82, 29.62] & 27.40* [26.13, 28.67] & 27.72* [25.26, 30.18] \\
Ministral & 21.72* [20.83, 22.61] & 21.28 [17.86, 24.70] & 23.64* [22.01, 25.27] \\
Gemma-2 & 28.92* [25.74, 32.10] & 23.00* [19.48, 26.52] & 30.52* [28.69, 32.35] \\
GPT-4o & 29.64* [27.31, 31.97] & 29.32* [26.97, 31.67] & 28.76* [27.44, 30.08] \\
Claude-3.5 & 49.64* [47.42, 51.86] & 52.08* [48.24, 55.92] & 54.40* [53.02, 55.78] \\
Gemini-2.5 & 34.29* [28.58, 40.00] & 37.17* [34.64, 39.70] & 42.10* [39.70, 44.50] \\
\bottomrule
\end{tabular}
\end{table}

\section*{Llama-3.1 temperature sensitivity analysis}\label{appd:temperature_ablation}
We conducted experiments with Llama-3.1 8B at temperatures 0.0 (greedy decoding producing deterministic output) and 0.3 (low variance), comparing results with our primary analysis at temperature 0.7.

We generated 2,500 responses per temperature setting across five random seeds. 
At temperature 0.0, we observed minimal duplication (4 duplicates across 2,500 responses). 
Average response lengths were comparable across temperatures: 222 words at $T=0.7$ [218, 226], 229 
words at $T=0.0$ [223, 235], and 228 words at $T=0.3$ [223, 234]. 
We applied identical analysis pipelines with Bonferroni-corrected significance testing ($\alpha = 0.05 / 6 = 0.0083$, accounting for three classifiers × two temperature conditions per characteristic).

\begin{figure}[!ht]
  \centering
  \begin{subfigure}[t]{0.48\textwidth}
    \centering
    \includegraphics[width=\linewidth]{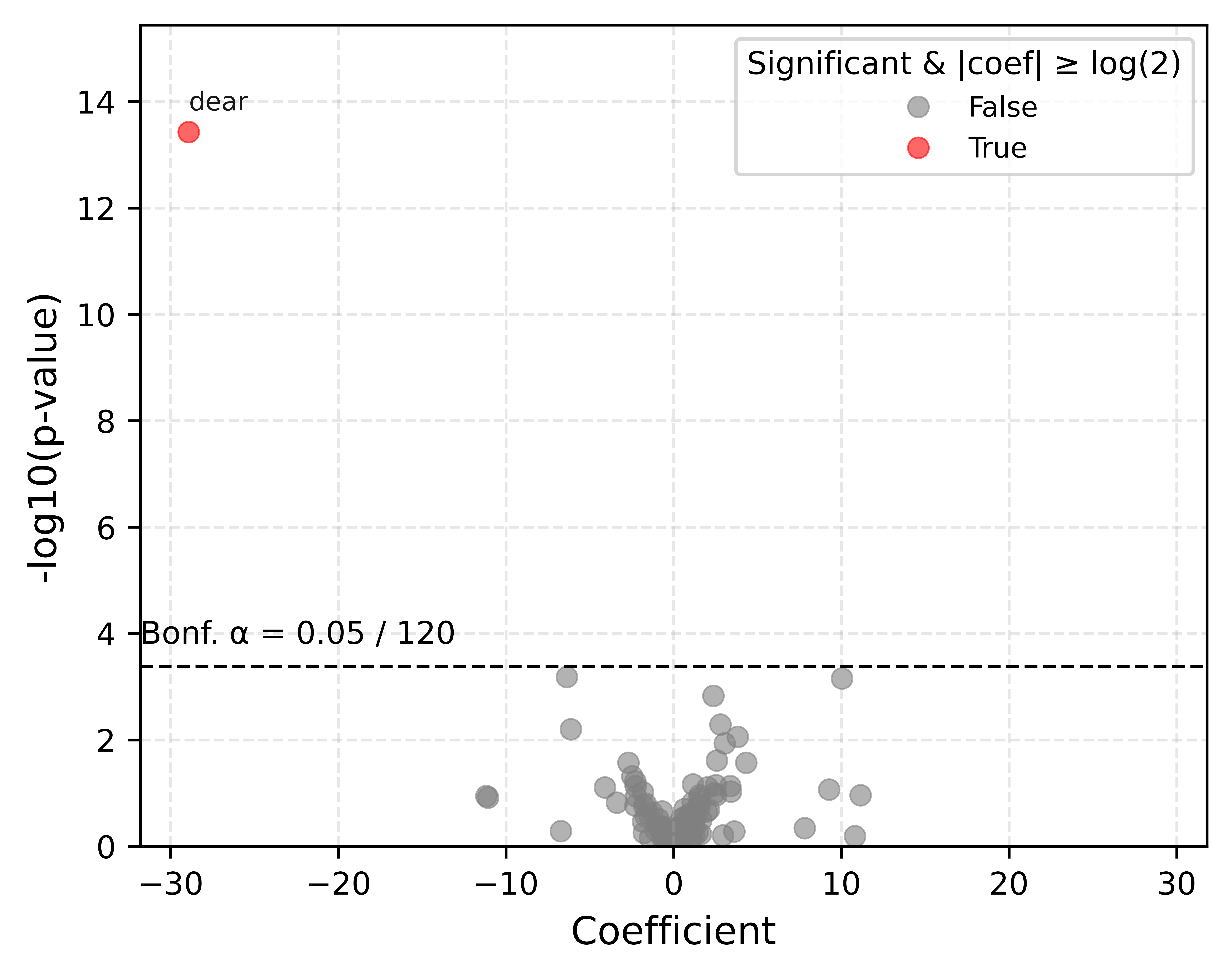}
    \caption{Temp = 0.0}\label{fig:llama_sex_temp0}
  \end{subfigure}\hfill
  \begin{subfigure}[t]{0.48\textwidth}
    \centering
    \includegraphics[width=\linewidth]{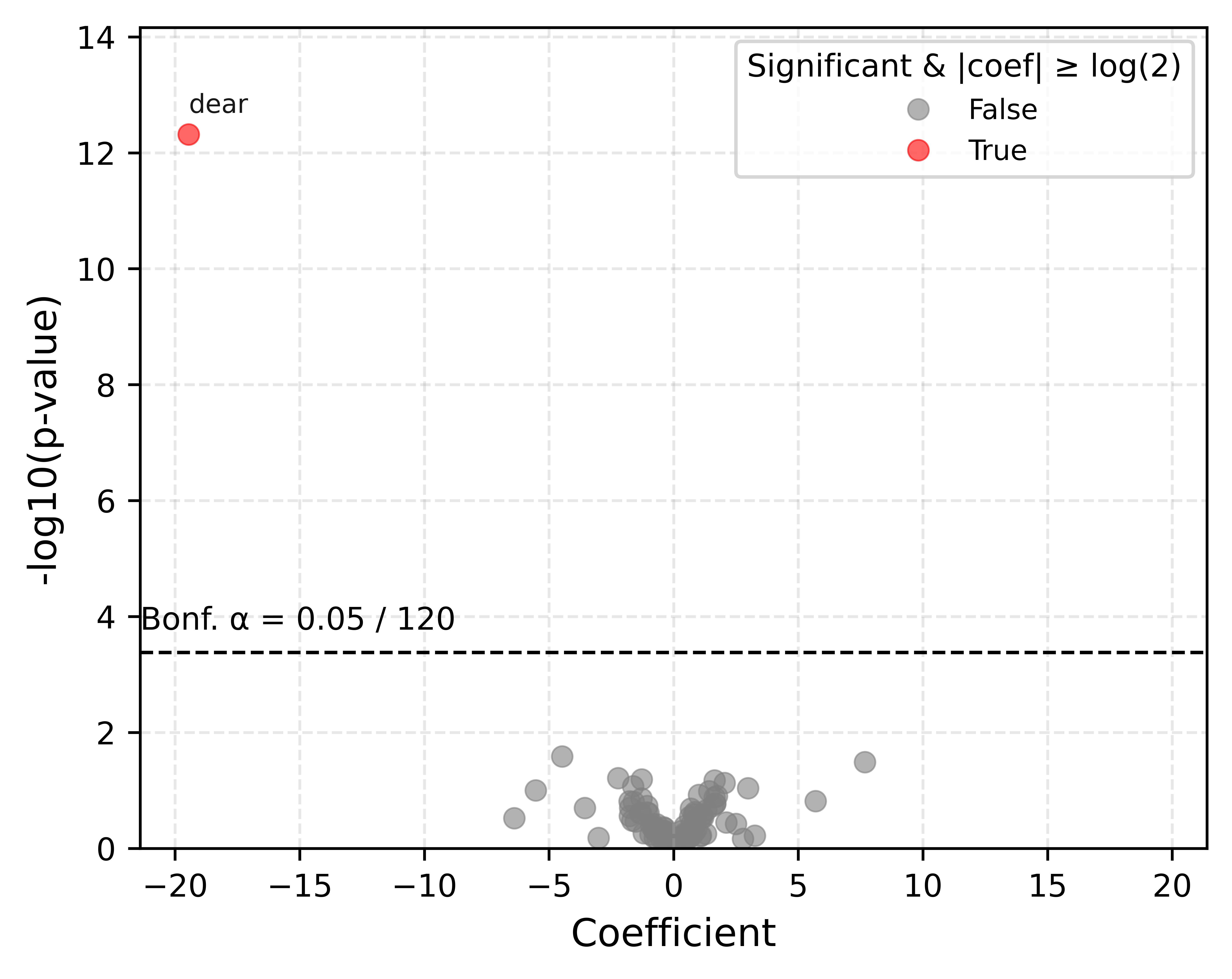}
    \caption{Temp = 0.3}\label{fig:llama_sex_temp03}
  \end{subfigure}
  \caption{Volcano plots for sex classification at temperatures 0.0 and 0.3. Each plot shows feature importance (x-axis: log-odds coefficients) versus statistical significance (y-axis: $-\log_{10}(p)$). The salutation ``dear'' remains the only statistically significant feature after correction across both temperature settings, consistent with findings at $T=0.7$.}
  \label{fig:llama_sex_ablation}
\end{figure}

Fairness patterns remain qualitatively consistent across temperatures, confirming that our conclusions are robust to the choice of sampling temperature. 
For sex, the same linguistic feature (``dear'') drives classification across all temperatures.
At $T=0.7$, only logistic regression reached significance (+5.00\%*), whereas at $T=0.0$ and $T=0.3$, all three classifiers achieved significance (Table \ref{tbl:ablation_sex}), with XGBoost showing the strongest performance (+8.00\%* and +8.28\%* respectively).
Volcano plot analysis (Figure \ref{fig:llama_sex_ablation}) confirms that ``dear'' remains the sole statistically significant feature at $T=0.0$ and $T=0.3$, replicating the pattern observed at $T=0.7$.
This consistency demonstrates that the sex-related pattern reflects a genuine systematic preference in salutation style rather than temperature-dependent artefact.

For race/ethnicity, the marginally significant pattern observed at $T=0.7$ (LogReg +1.93\%* margin) disappeared entirely at both $T=0.0$ (+1.01\%) and $T=0.3$ (-0.19\%), with no classifiers reaching significance (Table \ref{tbl:ablation_race_ethnicity}). 
This absence of replication at lower temperatures confirms that the $T=0.7$ result represented a marginal statistical fluctuation rather than genuine racial/ethnic bias, strengthening our conclusion that Llama-3.1 maintains demographic neutrality across race and ethnicity groups.

Patron type classification remained significant across all temperatures (Table \ref{tbl:ablation_patron_type}), exhibiting the same pattern of institutional role accommodation with amplified effect sizes at lower temperatures: $T=0.0$ showed margins of +20.69\%* to +35.49\%*, compared to +15.13\%* to +19.21\%* at $T=0.3$ and +10.05\%* to +11.05\%* at $T=0.7$.
The consistent significance across all temperature settings confirms that institutional role accommodation is not sensitive to temperature choice.

\begin{table}[!h]
\centering
\caption{Classification performance margins over random chance (16.67\%) for predicting user race/ethnicity from Llama-3.1 responses. Asterisks (*) indicate significance after Bonferroni correction ($\alpha = 0.0083$ for ablation; $\alpha = 0.0028$ for primary analysis).}
\label{tbl:ablation_race_ethnicity}
\small
\begin{tabular}{lccc}
\toprule
\textbf{Temperature} & \textbf{LogReg} & \textbf{MLP} & \textbf{XGBoost} \\
\midrule
0.0 & +1.01 & +0.21 & +1.01 \\
0.3 & -0.19 & -1.19 & -0.03 \\
0.7 (primary) & +1.93* & +0.97 & +1.61 \\
\bottomrule
\end{tabular}
\end{table}

\begin{table}[!h]
\centering
\caption{Classification performance margins over random chance (50.00\%) for predicting user sex from Llama-3.1 responses. Asterisks (*) indicate significance after Bonferroni correction ($\alpha = 0.0083$ for ablation; $\alpha = 0.0028$ for primary analysis).}
\label{tbl:ablation_sex}
\small
\begin{tabular}{lccc}
\toprule
\textbf{Temperature} & \textbf{LogReg} & \textbf{MLP} & \textbf{XGBoost} \\
\midrule
0.0 & +5.52* & +4.68* & +8.00* \\
0.3 & +4.96* & +3.68* & +8.28* \\
0.7 (primary) & +5.00* & +1.96 & +2.32 \\
\bottomrule
\end{tabular}
\end{table}

\begin{table}[!h]
\centering
\caption{Classification performance margins over random chance (16.67\%) for predicting user patron type from Llama-3.1 responses. All results significant after Bonferroni correction ($\alpha = 0.0083$ for ablation; $\alpha = 0.0028$ for primary analysis).}
\label{tbl:ablation_patron_type}
\small
\begin{tabular}{lccc}
\toprule
\textbf{Temperature} & \textbf{LogReg} & \textbf{MLP} & \textbf{XGBoost} \\
\midrule
0.0 & +20.69* & +33.77* & +35.49* \\
0.3 & +15.13* & +15.29* & +19.21* \\
0.7 (primary) & +10.05* & +10.73* & +11.05* \\
\bottomrule
\end{tabular}
\end{table}

\section*{Radar plot coefficients}\label{appd:radar_data}

\begin{table}[!ht]
\centering
\caption{Significant logistic regression coefficients for consensus features across LLMs and patron types. Values represent log-odds relative to undergraduate baseline.}
\label{tbl:radar_coefficients_summary}
\small
\begin{tabular}{llrrrrr}
\toprule
\textbf{Feature} & \textbf{Model} & \textbf{Graduate} & \textbf{Faculty} & \textbf{Staff} & \textbf{Alumni} & \textbf{Outside} \\
\midrule
\multirow{4}{*}{\textbf{Thank}} 
& Llama-3.1 & +25.3 & +31.8 & +39.3 & +20.6 & +29.3 \\
& Ministral & -5.9 & -12.0 & -6.6 & -10.0 & -10.6 \\
& Gemini-2.5 & +32.0 & +12.5 & +28.9 & +21.3 & +1.8 \\
\midrule
\multirow{5}{*}{\textbf{Dear}} 
& Llama-3.1 & -11.1 & -17.0 & -20.8 & -12.7 & -21.5 \\
& Gemma-2 & +2.1 & -4.0 & -5.7 & -4.0 & -12.4 \\
& GPT-4o & +53.6 & +44.9 & -28.7 & +9.4 & -35.9 \\
& Claude-3.5 & --- & -9.0 & -43.8 & -25.8 & -34.4 \\
& Gemini-2.5 & -15.2 & -60.9 & -51.8 & -55.5 & -52.4 \\
\midrule
\multirow{6}{*}{\textbf{Research}} 
& Llama-3.1 & +1.7 & +4.7 & --- & --- & +2.6 \\
& Ministral & +5.4 & +8.7 & +3.6 & --- & +8.8 \\
& Gemma-2 & +7.5 & +9.8 & +1.2 & +1.9 & +5.8 \\
& GPT-4o & +5.4 & +7.6 & -2.4 & +2.3 & +4.4 \\
& Claude-3.5 & +5.7 & +7.1 & +2.4 & +2.6 & +6.0 \\
& Gemini-2.5 & +17.6 & +32.6 & +7.5 & +6.0 & +24.2 \\
\bottomrule
\end{tabular}
\end{table}

\end{document}